\documentclass{article}

\usepackage[english]{babel}
\usepackage[superscript,biblabel]{cite}
\usepackage{authblk}
\usepackage[a4paper,top=3cm,bottom=2cm,left=2cm,right=2cm,marginparwidth=1.75cm]{geometry}
\usepackage{setspace}
\usepackage{amsmath}
\usepackage{xurl}
\usepackage{graphicx}

\usepackage[colorlinks=true, allcolors=blue]{hyperref}
\date{}

\graphicspath{{Figures/}}
 
 \usepackage{adjustbox}
 \usepackage{array}
\newcolumntype{P}[1]{>{\centering\arraybackslash}p{#1}}

\title{Industrial Data Science for Batch Manufacturing Processes}
\author[a]{I. Imanol Arzac}
\author[b]{Mattia Vallerio}
\author[b]{Carlos Perez-Galvan}
\author[b,c,*]{Francisco J. Navarro-Brull}
\affil[a]{Department of Chemical Engineering, KU Leuven}
\affil[b]{SOLVAY SA, Belgium}
\affil[c]{Department of Chemical Engineering, Imperial College London}
\affil[*]{Corresponding contributor. Email: francisco.navarro@solvay.com, f.navarro@imperial.ac.uk}

\begin{document}
\maketitle

\newpage 

% ABSTRACT HERE
\begin{abstract}
Batch processes show several sources of variability, from raw materials' properties to initial and evolving conditions that change during the different events in the manufacturing process. In this chapter, we will illustrate with an industrial example how to use machine learning to reduce this apparent excess of data while maintaining the relevant information for process engineers. Two common use cases will be presented: 1) AutoML analysis to quickly find correlations in batch process data, and 2) trajectory analysis to monitor and identify anomalous batches leading to process control improvements.
\end{abstract}
\newpage

\tableofcontents

\newpage
\section{Introduction} \label{introduction}

Batch processes are the oldest form of manufacturing \cite{levenspiel1998chemical, levenspiel2013chemical}, ---follow this recipe under these conditions for this amount of time, and you should obtain adequate products. Contrary to continuous processes ---where steady-state conditions facilitate the use of monitoring, control, and optimization methods---, batch operations show inherent variability through the various manufacturing steps followed in a recipe. Advances in automation and control for batch processes have enabled more consistent quality and productivity (e.g. trajectory control, MPC).\cite{Kourti2005,Lin2009,Kanavalau2019,Srinivasan2003, macgregor_MPC_batch,rawlings2017model,vallerio2014tuning }

In recent years, process industries have invested in machine learning teams, software, and infrastructure due to the promise of data-driven applications in manufacturing \cite{Beck2016May, McKinsey2015, McKinsey_APC, Chiang2022, SANSANA2021_dow_review_Ind_40, DSCCM20, Zhang2022, Mck_2022_aug_trends}. Unlike big tech companies ---on which online recommendation systems allow quick iterations by trial-and-error---, manufacturing industries must deal with the safety of such recommendations and the inevitable challenges imposed by the physicochemical, engineering, and operation constraints \cite{Piccione2019Jul,Clarke2016Nov, SHANG20191010}. Similarly, IT challenges evolve around moving data to the cloud while, in reality, industrial data is mostly generated and consumed locally \cite{McKinsey2019}. And yet, the affordability of modern hardware and software (both private and open-source) can reduce obstacles and analysis time, in case the required knowledge is available to choose the right set of tools and to apply them to the process in hand \cite{Chiang2022,SANSANA2021_dow_review_Ind_40}. 

In this work, we will review the challenges linked to analyzing industrial batch data and how to obtain valuable insights using machine learning ---both regression (supervised learning) and classification methods (unsupervised learning). 

\paragraph{Who's this book chapter for?}

The analysis and discussion of this book will illustrate how data-driven methods can allow process engineers to quickly monitor and troubleshoot industrial batch processes in a manufacturing environment. Therefore, our focus will be to present the intuition behind machine learning methods and their industrial applications pragmatically. Most of the analysis will be done with point-and-click commercial software (JMP Pro, SAS Institute Inc \cite{jones2011jmp}). However, all machine learning methods explained here are also available in open-source packages: Python (scikit-learn\cite{scikitlearn}, pycaret\cite{PyCaret}, scikit-fda\cite{Ramos-Carreno_GAA-UAM_scikit-fda_Functional_Data_2019}, pyphi\cite{pyphi_salvador}) and R  (caret\cite{caret_kuhn2008building}, h2o\cite{h2o_R_package}, mlr\cite{R_mlr_package}, fdapace\cite{yao_fang_2005_FDA}); as well as other commercial software (e.g. SEEQ\cite{peterson2018improving_SEEQ}, Trendminer\cite{TrendMiner_web}, Aspen ProMV\cite{batch_ProMV_proceeding, Ghosh_ProMV}, and SIMCA\cite{UNDEY2009}). 

The interested reader can look deeper into machine learning foundations by following the provided references or the technical documentation of the cited software and packages.

\newpage
\section{Industrial Batch Data and Analytics}  \label{Ind_batch_data_analyics}

\subsection{Batch data}

\begin{figure}[!htb]%[!htb]
    \centering
        \includegraphics[scale=0.55]{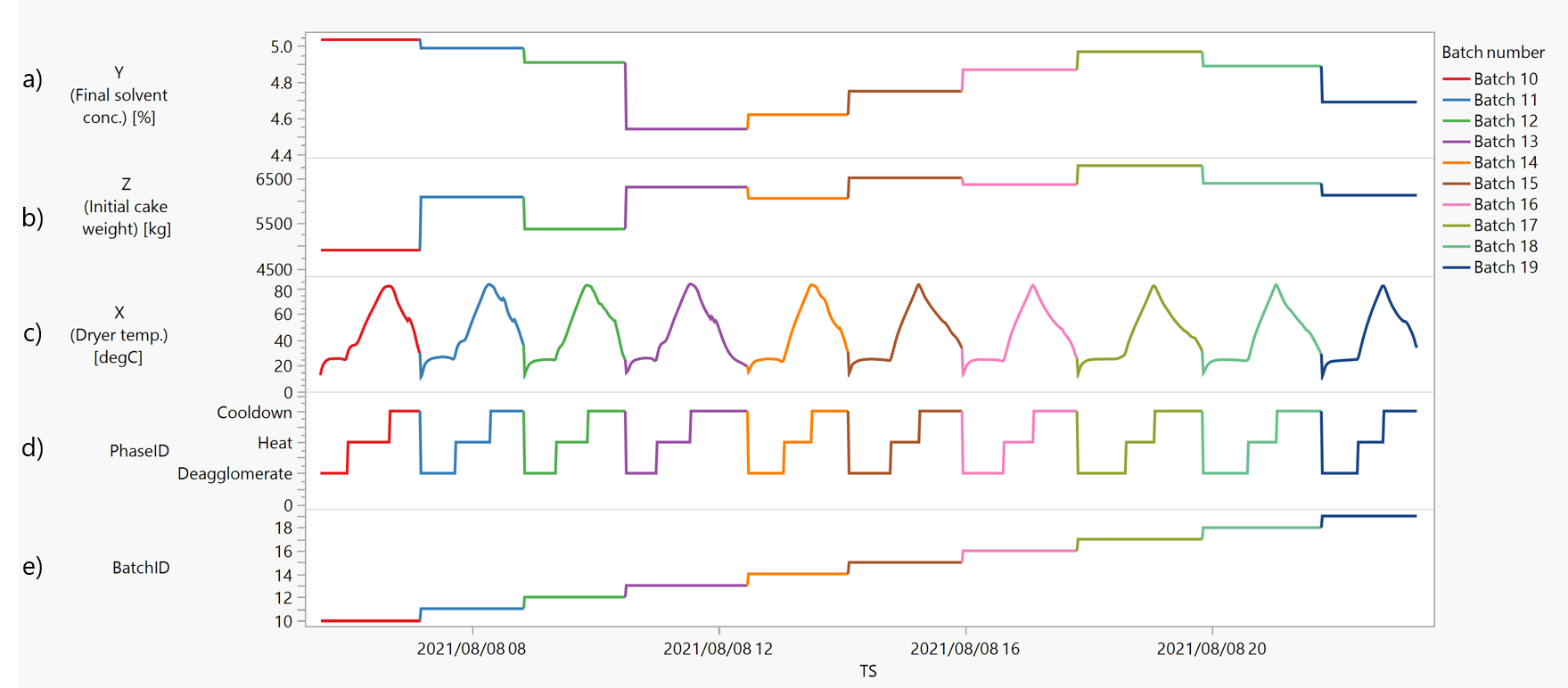}
    \caption{Batch processes have a combination of different data types. Single measured values such as concentration, quantity, or quality for end products (a, named usually as target or Y variables) and initial conditions (b, defined as Z variables) that include raw materials. Time-varying data coming from sensors such as reactor temperature (c) which evolves through the process showing trends or trajectories (defined as inputs or X variables); and event data (d and e) which determines in which phase and in what batch the manufacturing process was.}
    \label{fig:batch_data_historian}
\end{figure}

Utilizing all the data from batch processes represents a challenge, as the model output can be a single lab measurement (e.g. predicting product quality). In contrast, model inputs range from raw materials properties to initial and evolving conditions (e.g. pressure and temperature) that were measured during the different phases of the batch (see Fig. \ref{fig:batch_data_historian}).

In this chapter, we will describe two common workflows to monitor and troubleshoot an industrial drying/reaction process from the literature \cite{garcia_munoz_batch_2003, garcia_munoz_batch_2004}. The reactor is charged with a variable amount of wet cake that evaporates and recollects a solvent material in an external tank. There are three distinct phases in the batch dataset (see Fig. \ref{fig:batch_data_historian}):
\begin{enumerate}
\item Deagglomeration phase: where the cake reacts at low agitation speed while the solvent is collected 
\item Heating phase: where the temperature of the cake is increased until its set-point
\item Cooling phase: where the batch temperature is reduced before unloading.
\end{enumerate}
During the drying process, there are important structural and chemical reactions that also take place. At the beginning of the batch, a variable amount of cake is measured with an unknown amount of solvent (Z). The operator can modify both the dryer temperature and agitation speed.  At the end of the batch, a sample is taken and sent to the laboratory for quality control. The percentage of remaining solvent (Y) is an important parameter determining whether the batch is in-spec or out-of-spec. A total of 10 tags (recorded information from the sensors or control) can be downloaded from the author's python library\cite{pyphi_salvador}. More information about the batch can be found in their published work\cite{garcia_munoz_batch_2003, garcia_munoz_batch_2004}.

In manufacturing, additional event data is commonly present. For example, what grades (products), campaigns, and lots the reactor was producing and whether it was in cleaning and maintenance operations. In addition, if a manufacturing standards such as ISA-88 and ISA-95 are implemented (see Section  \ref{Chall_opportunities}), a global hierarchy describes where the batch process is in the plant. This hierarchy is based on an asset tree structure that assigns sensors (called tags) to their respective control modules, equipment, unit, process section, production plant and site.

In the literature,  \cite{Wold2009, garcia_munoz_batch_2003, garcia_munoz_batch_2004, Qin2012} different approaches have emerged to effectively reduce this apparent excess of data (dimensionality) while maintaining the information to detect and understand anomalies or use predictions for control \cite{Spooner2018a, Spooner2018b, zuecco2021backstepping, ZUECCO_BASF_2020_batch, Gonzalez-Martinez2011, Spooner2017}.

In the following sections, we will address the main data steps when analyzing batch data.

\subsection{Batch data alignment}

\begin{figure}[!htb]%[!htb]
    \centering
        \includegraphics[scale=0.65]{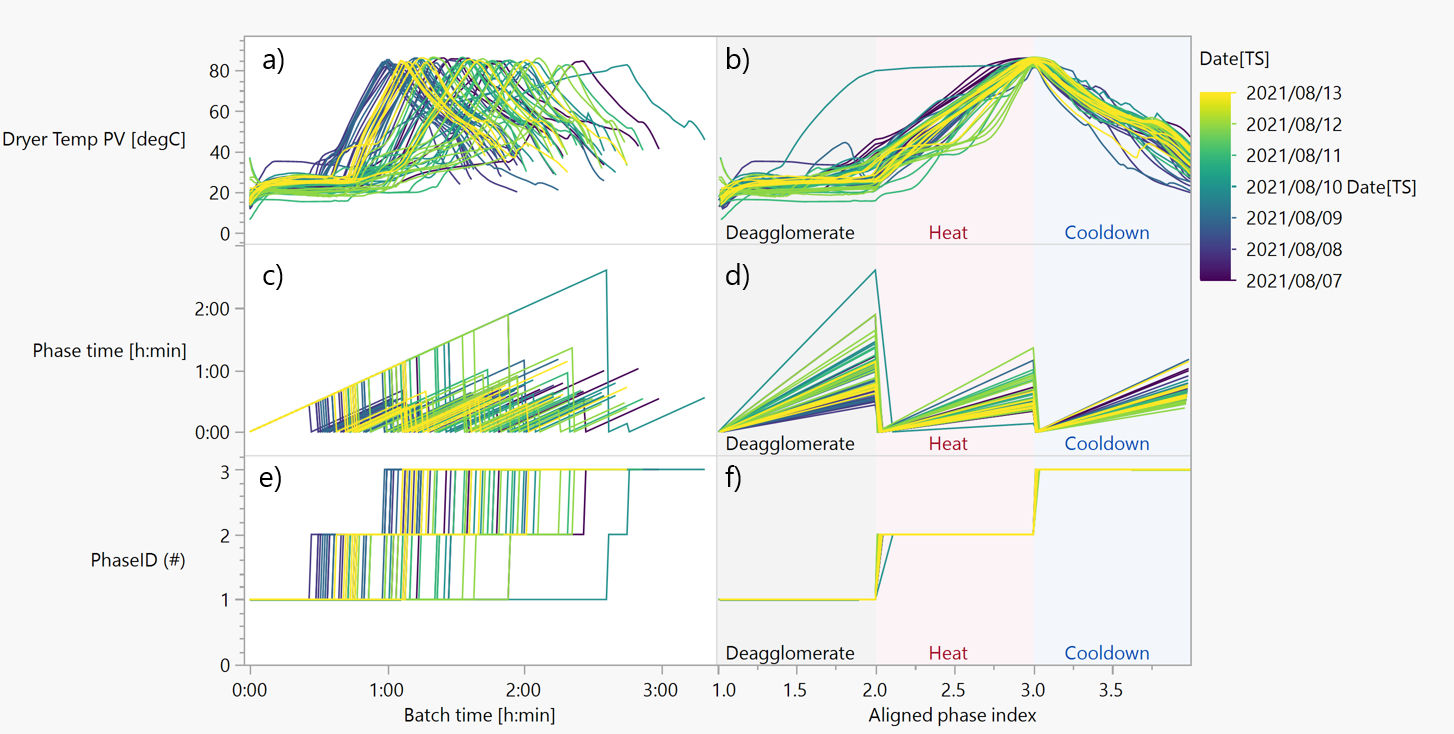}
    \caption{Overlay of a batch drying process \cite{garcia_munoz_batch_2003, garcia_munoz_batch_2004}, with the temperature profile (a) showing different durations (c) for its distinct phases (d) colored by date. A common challenge in manufacturing is that that the duration of each phase varies batch-to-batch. To aid the visualization (b), a simple alignment using the automation triggers (in this case, phase start and end) can be done (d,f). Notice that this simple alignment technique\cite{GarciaMunoz2011} allows the finding of anomalous batches by visual inspection (or density-based analysis, not shown).}
    \label{fig:Overlay_batch_data_aligned}
\end{figure}

The drying process is a good illustration of how the duration can vary in batch manufacturing processes (see Fig. \ref{fig:Overlay_batch_data_aligned}). In the dryer dataset, the main source of variability comes from the batch-to-batch variation in the amount of product loaded and its solvent content.

In general, phases can take longer due to several process perturbations that affect  kinetics (e.g. catalyst deactivation), raw material variability (due to changes in quality or quantity), reduction of heating/cooling capacity, or, simply, variability due to maintenance issues or scheduling decisions. This inherent complexity will interfere with the variability expected through the steps in one of several procedures (automated or manual) of a batch manufacturing process.

To align batch data with different duration, several techniques exist \cite{Wold2009, zuecco2021backstepping,GarciaMunoz2011,Kassidas1998,Brunner2021}. However, as pointed out in the literature \cite{GarciaMunoz2011}, automation triggers can be used to automatically align the batches when first-principle (e.g. conversion of the reaction or amount of water evaporated instead of time) is not available. When information to align batches is not measured nor known, Dynamic Time Warping (DTW) techniques align batch trajectories statistically \cite{Wold2009, zuecco2021backstepping, Spooner2017,Spooner2018a,Spooner2018b,Zhang2013,Kassidas1998,Ramaker2003,Gins2012,Gonzalez-Martinez2011,GonzalezMartinez2014}. 

Although DTW can be used to classify anomalous batches or to identify correlating parameters \cite{Spooner2018a, Spooner2018b, zuecco2021backstepping, Gonzalez-Martinez2011, Spooner2017}, aligning batch data using automation information has two main benefits:
\begin{itemize}
\item Parameter-free as it does not require a batch reference nor other tuning parameters
\item Straight-forward to apply as it involves only a calculation of a new time-index
\end{itemize}

\newpage

\begin{figure}[!htb]%[!htb]
    \centering
        \includegraphics[scale=0.7]{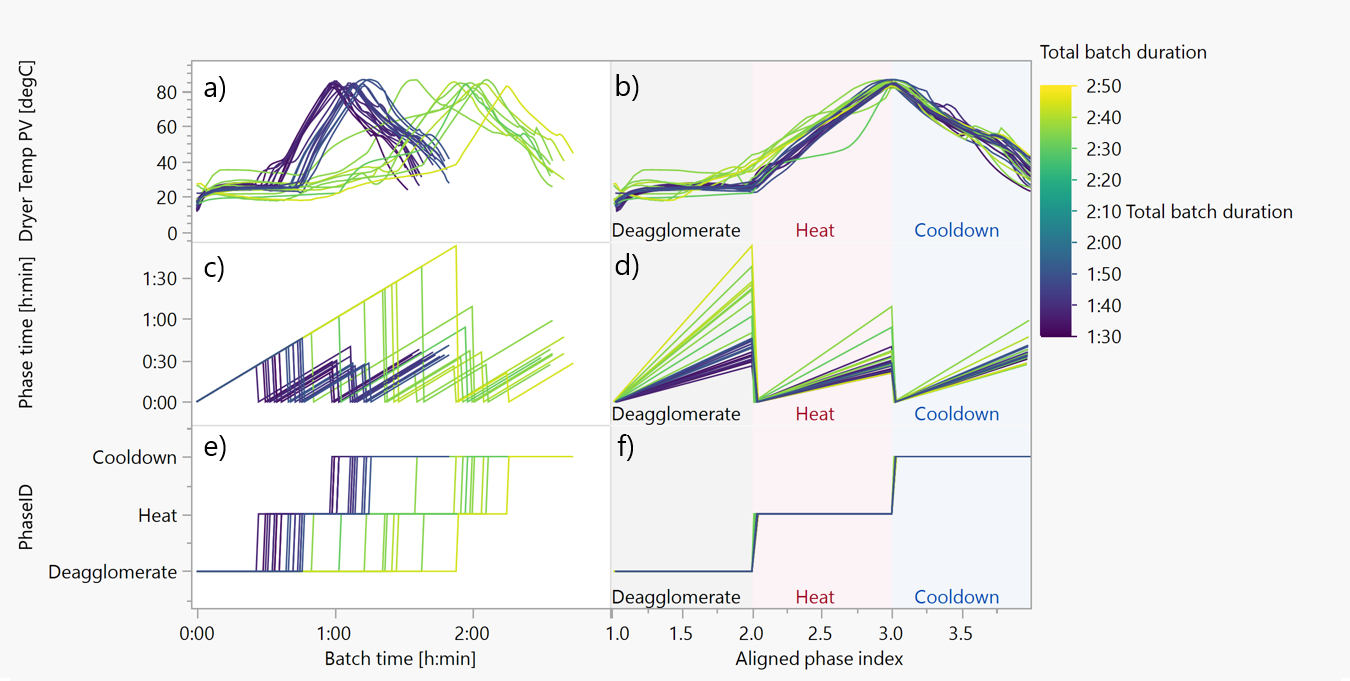}
    \caption{Overlay of the drying temperature profile (a) and phases (c, e) for a selection of historized batches colored by total batch duration instead of the batch date. When batches are aligned (b) between  phases (f), a higher duration to complete the same  phase will correspond to a steeper slope (d) in the visualization.}
    \label{fig:Overlay_duration_batch_data_aligned}
\end{figure}

For the interested reader, a comparison of batch data alignment algorithms is provided in Annex \ref{app:batch data alignment}, with a special focus on Dynamic Time Warping (DTW) techniques.

Once aligned, patterns start to be recognized. The duration of each phase may convey important information that should be kept in the analysis and the visualization (see Fig. \ref{fig:Overlay_duration_batch_data_aligned}).

\subsection{Batch data analytics}

\subsubsection{Summary statistics (a.k.a landmarks, fingerprints)}

\begin{figure}[!htb]%[!htb]
    \centering
        \includegraphics[scale=0.7]{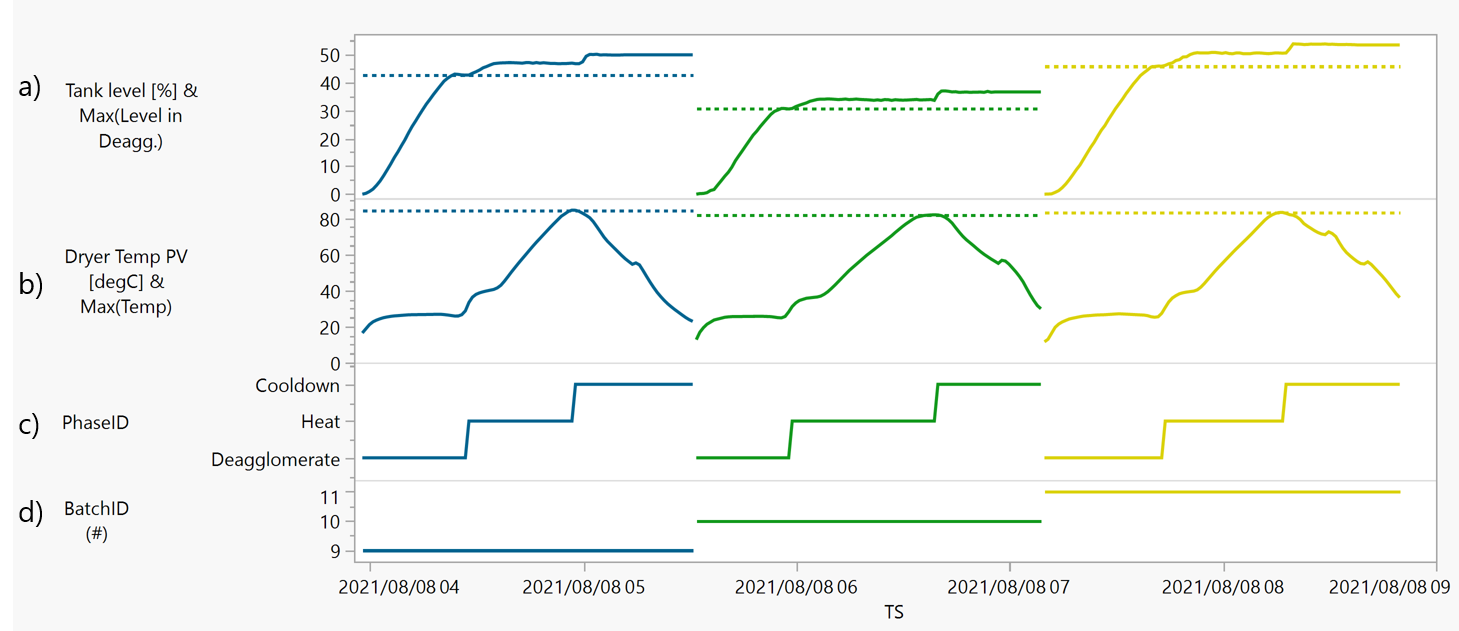}
    \caption{Model inputs for batch processes can be generated by summarizing the information into landmark points or fingerprints \cite{Wold2009}. Here, the tank level at the end of the deagglomerate phase (a) and the maximum temperature reached during the drying process (b) can be used to capture information in specific phases (c) and summarize it per batch (d). These single values are then used for either correlation analysis or statistical process control monitoring.}
    \label{fig:fingerprint}
\end{figure}

A common approach to deal with batch data with different time lengths is to summarize each batch using statistics and process knowledge (e.g. peak temperature or its average rate of change during the reaction phase). In the literature, these are known as landmark points or fingerprints (see Fig. \ref{fig:fingerprint}), but it assumes that subject matter experts (SMEs) know \textit{a-priori} the important features to generate. Generalizing this approach, one could calculate common statistics (average, max, min, range, std, first, last, or their robust equivalent), for every sensor, its derivative or integral, during every phase, for every batch and grade (product). In machine learning, this is known as feature engineering and is often combined with algorithms to keep the best predictors only (e.g., feature selection). An example of how to use ML to both generate and select a subset of sensor data will be shown in Section \ref{ML_applications}.%

As pointed out in the literature, the landmark approach is conceptually and technically the simplest form of analyzing batch data \cite{Wold2009}. Notice that the calculation of these features does not require alignment of data and it can be done directly in the historians (industrial databases that record sensor data in the plant). These are essentially group-by operations that can be easily queried, pre-calculated and stored in aggregated tables. The practical advantages of this method cannot be ignored, as it requires little implementation effort, avoids the need to extract all the sensor data to start the analysis and reduces impact of the system dynamics.

\subsubsection{Functional or shape analysis (FPCA)}

Instead of summarizing the sensor data in statistical calculations that can aggregate and dilute important information, the whole trajectory can be used instead. This is especially necessary if the batch-to-batch variability is minimal \cite{Wold2009}, or if a more detailed analysis for a subset of sensors is important to understand and explore further.

A common approach that has been widely used in the industry, is to manipulate the data table, so every sample point recorded in time by the sensor (rows) is used as a variable (columns).\cite{Wold2009, garcia_munoz_batch_2003, garcia_munoz_batch_2004, GarciaMunoz2011,Gins2016,MacGregor2012,MacGregor1994,Spooner2018b,Wold2009,Wold1987,Ramos2021,Nomikos1995,Nomikos1994,Rendall2019,Ramaker2002,Chiang2022,Ramaker2006, MACGREGOR2015_big_data} Batches with different lengths need to be re-sampled using interpolation techniques, so every sensor has the same number of points per batch (again, data points will become columns in the new table). This data manipulation technique allows the direct use of Principal Component Analysis for batch anomaly detection, and multivariate correlation analysis (in this case using Partial Least Squares). The reader is referred to the following book chapter\cite{Wold2009} for a comprehensive discussion and their application to several batch processes in the industry (including the dryer dataset used in this work).

\paragraph{What's PCA used for?}
Before advancing the discussion, let's briefly explain the main intuition behind PCA in process engineering. Principal Component Analysis is the go-to approach when trying to find latent variables in a manufacturing process \cite{KevinDunn2010_PID,Dong2018} ---that is, to reduce redundant sensors into the main drivers of a process. For example, if several sensors are measuring the fluid temperature in a pipe, a common engineering approach will be using a weighted average while controlling that no deviations occur which can indicate sensor failure. PCA is used for this exact purpose, and it can find all redundant sensors that are linearly correlated (also called co-linear). PCA then reduces them into independent latent variables (e.g. weighted averages of pressure, temperature, etc.), which are also called principal components of a dataset. Recently, new techniques such as UMAP\cite{mcinnes2020umap}, have shown to outperform PCA when applied to process data\cite{Joswiak2019}.

\paragraph{What's the 'functional' part in FPCA?}

\begin{figure}[!htb]%[!htb]
    \centering
        \includegraphics[scale=0.85]{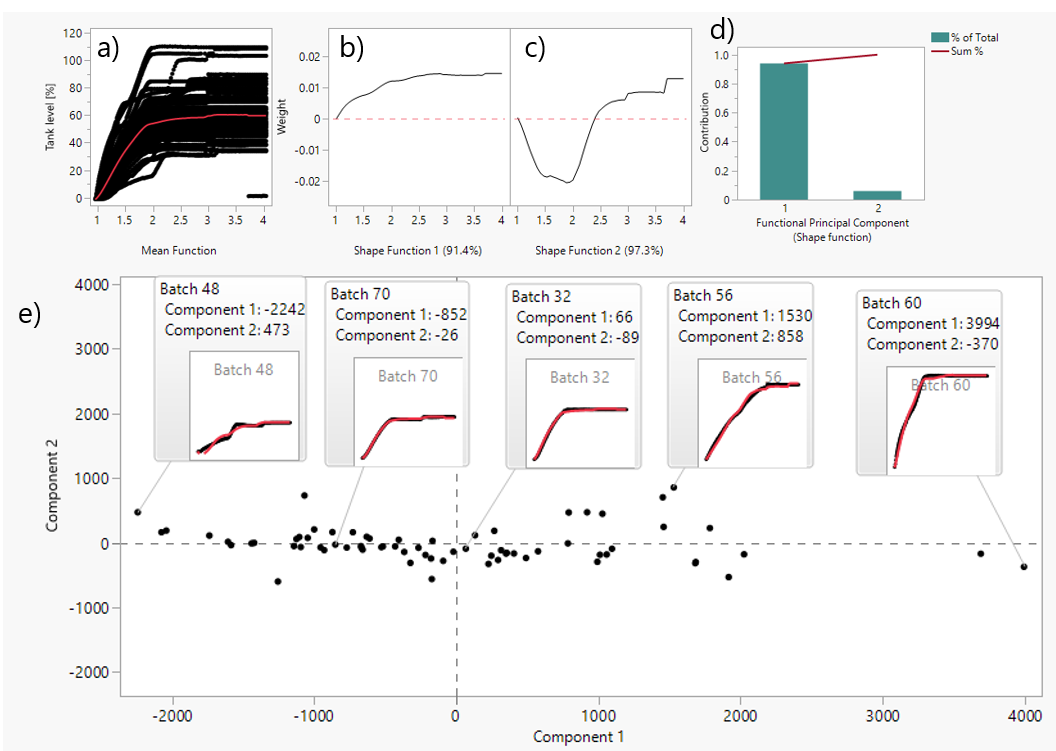}
    \caption{Contrary to landmarks or fingerprints, Functional PCA can capture and decompose the whole trajectories seen in batch processes. In the image, batch curves can be described again using: (a) the average trajectory and a combination of characteristic and independent trajectories (b, c; called shape functions) which have different importance (d) as the major source of variability in this example is the height reached by the level, for example. Using these shape functions, new coordinate variables (e) can be used to detect anomalous batches (e.g. batch number 60) or cluster them into different groups.}
    \label{fig:FPCA_level}
\end{figure}

\begin{figure}[!htb]%[!htb]
    \centering
        \includegraphics[scale=0.7]{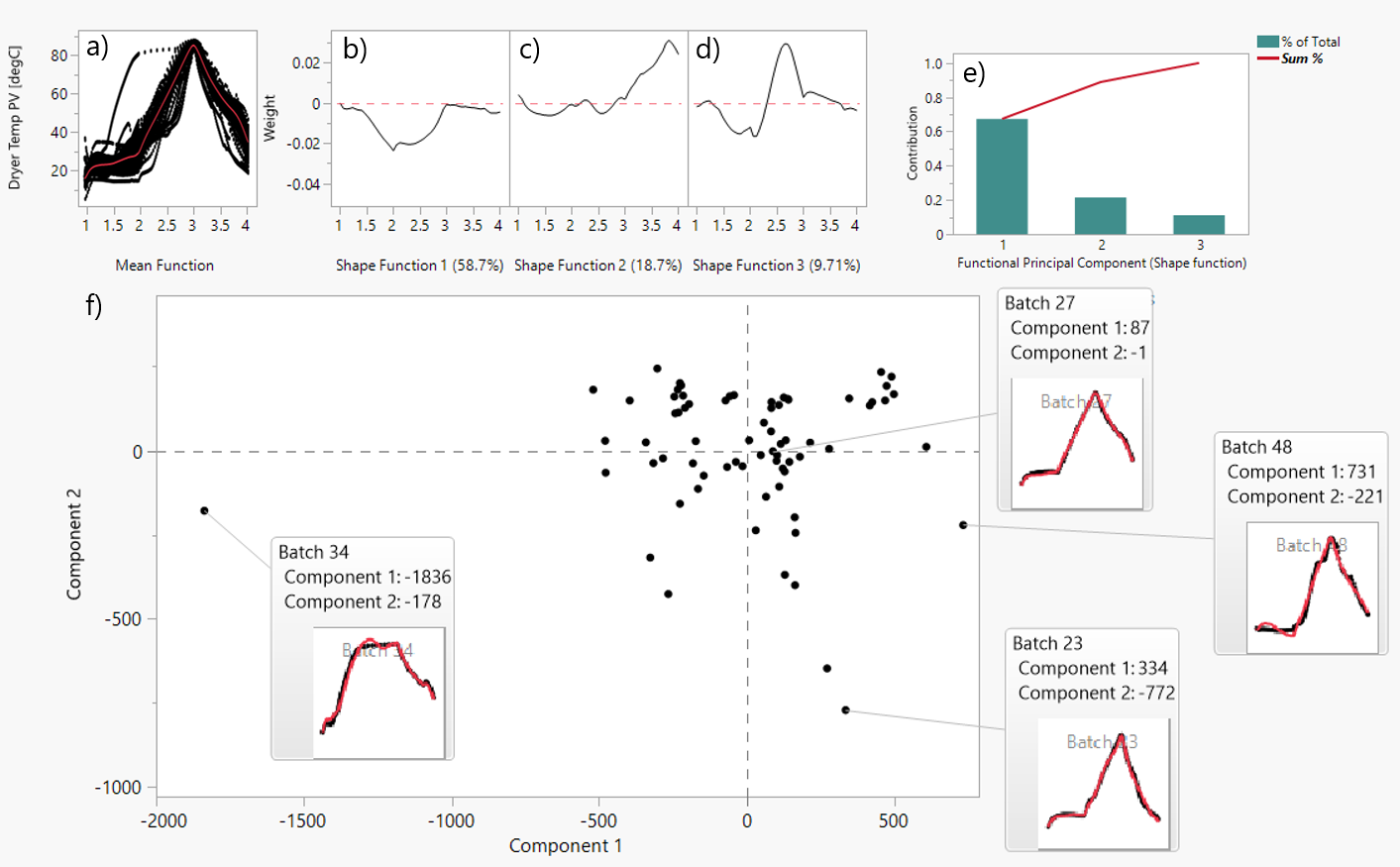}
    \caption{FPCA results for the temperature profiles in the dryer. The average trajectory is first calculated by aligning the batches phase by phase (see red line and time index in a). Then a combination of characteristic and independent trajectories (b, c, and d; called shape functions) are obtained and ordered by contribution to the total variability (e). Using the average and these shape functions (intrinsic to all batch data), new coordinate variables (f) can be used to describe each batch individually. The new coordinates and scores can identify anomalous batches (e.g. batch number 34 in f) or be used for clustering (not shown). The differences between the recalculated trajectories and the original data (red and black curves in f, respectively) show how accurate the FPCA analysis was at decomposing and recomposing batch data.}
    \label{fig:FPCA_temperature}
\end{figure}

Functional Principal Components Analysis (FPCA) is an extension of PCA used when the redundant information in data is represented as curves or trajectories \cite{silverman2002_applied_FPCA, kokoszka2017introduction_FPCA, srivastava2016functional}. Concretely for batch processes, FPCA captures the main sources of variation between multiple trends of a dryer level (see Fig. \ref{fig:FPCA_level}) or temperature (see Fig. \ref{fig:FPCA_temperature})  as a "function of batch time". FPCA is part of a field called Functional Data Analysis, which applications extend to other industrial domains: HPLC data (function of analysis time), spectroscopy data (function of wavelength), vibration (function of frequency), battery degradation\cite{Ullah_battery_FPCA_2022}(function of cycle time).\cite{Liu2020,Wang2015}

With FPCA, these trends or trajectories are summarized with a mean curve and a series of “shape functions” (called eigenfunctions). These have associated weights (called loadings or eigenvalues) ordered by total contribution ---again, each shape tries to summarize the variability seen in latent trajectories\cite{ramsay2006}.

If the batch data is pivoted and resampled as explained at the beginning of this section, FPCA insights might seem analogous to standard PCA. In the end, both seek to reduce the data into a smaller number of components describing as much information in the data as possible.
However, FPCA has the advantage of finding a set of component shapes that explain the maximum variation in the observed data. These shape components can be interpreted as distinctive features seen in the process for some batches. For example, a different rate in the deagglomeration phase for the level, or a temperature “shoulder” at a certain point in the heating phase.

Ultimately, each original batch trend can be reconstituted with the mean trajectory and a linear combination of these inherent weighted shapes. The specific amount of each shape function needed is called the FPC score, which varies from batch to batch. Not all the shape functions have the same contribution or importance, hence the weight (load or eigenvalue) must always be considered when looking at score plots (Fig. \ref{fig:FPCA_level}e,  \ref{fig:FPCA_temperature}f).  

 \paragraph{FPCA data requirements}
The first step is to turn the discrete data of the sensor value at each timepoint for each batch into a continuous function. This is done by fitting smoothing models, such as splines. This means it is possible to use both dense (observations are on the same equally spaced grid of time points for all batches) and sparse (batches have different numbers of observations and are unequally spaced over time) functional data. Then a functional principal components analysis can be carried out.

FPCA also requires a pre-alignment of the data to properly capture the average and principal shapes. In the example, we pre-aligned the batch using phases captured from the automation layer. In this regard, DTW can also be used to pre-align the data (see Annex \ref{app:batch data alignment}) and classify anomalous batches and investigate parameters that affect batch duration. \cite{Spooner2018a, Spooner2018b, zuecco2021backstepping, Gonzalez-Martinez2011, Spooner2017,Kassidas1998,Ramaker2003,Zhang2013,Keogh2001}.
 
Similarly, the results from FPCA, especially the FPC scores, can be saved and used as summary features for further analysis (such as classification, clustering of batches and regression analysis\cite{morris2015functional_regression}). For example, in the score plots, we can already identify batches far from the rest (e.g. batch 34 in Fig. \ref{fig:FPCA_temperature}f). 

The following section will introduce several applications combining both landmarks and FPCA scores.

\newpage
\section{Industrial Machine Learning Applications} \label{ML_applications}

Traditionally, rules-of-thumb and empirical correlations served as the foundation for the design of chemical process equipment \cite{Piccione2019Jul}. This is the same goal of Machine Learning: the ability to predict an outcome using experimental data from similar processes (e.g. prediction of heat-transfer coefficients or what fluid dynamic regime a system will be in \cite{Mowbray2022_ML_review}).

In this section, we will showcase two main industrial applications for batch data using both supervised and unsupervised machine learning.

\subsection{Correlation analysis for batch processes (supervised learning)}

When the problem to solve involves a known and specific variable to analyze (Y, also called target or model output), the information is called to be 'labeled' and a supervised Machine Learning. The target can be numeric, such as data coming from a sensor, or categorical (e.g. in-spec or out-spec).

In chemical engineering, we are familiar with correlation analysis since we want to understand what input variables (X's) fit the output variables (Y's).

\begin{figure}[!htb]%[!htb]
    \centering
        \includegraphics[scale=0.7]{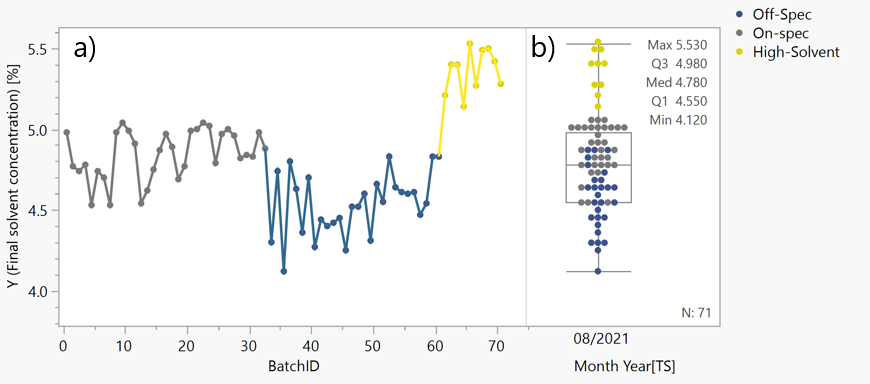}
    \caption{After drying, the solvent content is measured in the lab. Recently, it has shown variability over time (a), first being off-spec and then too high concerning the on-spec target (b). Using this batch-to-batch solvent content as a target (Y), a correlation analysis to reduce the number of sensors (tags) can be done.}
    \label{fig:Y_target}
\end{figure}

For example, the dryer dataset shows variability in one lab analysis taken at the end of the batch process. As shown in Fig. \ref{fig:Y_target}, the solvent content has decreased and then increased in recent batches, making them out-of-spec or too high according to past values.

The target variable (Y) is the most important variable to define, so it needs to be pre-processed adequately. The ML analysis will identify correlated input variables (called X's or predictors) that can explain the described variability. If there are known effects  (e.g. shutdowns, summer/winter seasonality, calibration issues, etc.) the best approach will be to remove them from the target, not the list of X's. 

Similarly, if there has been a major process change, we won't probably find the cause to be proportional (i.e. linearity of inputs and outputs does not hold under big perturbations). Next, we will show how to find non-linear correlations using a robust modeling approach.

\subsubsection{AutoML for batch processes}
AutoML aims to simplify the use of machine learning problems by automating complex analytical tasks often done manually. But how? The premise of ML is that there is an excess of data which allows fitting several models (training dataset), on which model parameters are selected avoiding overfitting (validation dataset) and finally assessed with unseen data (test dataset). As this assumption can fall short in industrial applications, a more realistic scope for autoML is quickly finding sensors correlated to the problem at hand.

\paragraph{Feature engineering} As we saw in Section \ref{Ind_batch_data_analyics}, one can automatically calculate several summary statistics of a batch process (named landmarks, fingerprints, or features in ML). Once all sets of features have been calculated per product, batch, and phase, these can be used as model inputs (X's) directly. Notice that these can be: 1) as granular as needed, summarizing statistics even per automation step (which are combined in phases); and 2) using all process knowledge available (e.g. square root of pressure will make a pressure drop linear with the flow rate). In the literature, a well-known software package for feature engineering in the context of process system engineering is ALAMO \cite{wilson2017alamo} and for reaction estimation parameter estimation there is RIPE\cite{Nannicini2021}. In the literature, those approaches that combine data with first principle are called hybrid or grey-box models.\cite{SANSANA2021_dow_review_Ind_40, Bikmukhametov2020Jul, macgregor_hybrid_PLS}. 

\paragraph{Feature selection (screening models)}

Given the high number of X's created in the feature engineering step, any ML model should be able to handle, in an efficient way, the following:
\begin{enumerate}
\item Non-linear behavior, as the relationships between inputs and outputs won't always be proportional to each other (e.g. reaction rate with respect to reaction temperature).
\item High similarity between inputs, as several summary statistics are calculated automatically and will show co-linearity.  
\item Noise, as there will be random perturbations affecting both our target and inputs coming from sensors or lab analysis.
\end{enumerate}

We recently reviewed\cite{Mowbray2022_ML_review} a simple workflow that satisfies the three conditions, specifically:

\begin{enumerate}
\item Decision tree models handle non-linearities as they can create an elaborate logic using if-then rules (e.g. if flow and temperature are under these conditions, then this specific response happens)
\item By sub-sampling the data (e.g. random subset of X's) and then creating multiple decision trees, one can combine all the models and calculate the average response and contribution of each predictor. This is the main idea behind one of the Random Forest algorithms  ---called bootstrap forest or predictor screening in JMP (SAS Institute).
\item If synthetic noise is added as a sensor, one can use the contribution of this predictor as a cut-off for variable selection.
\end{enumerate}

\begin{figure}[!htb]%[!htb]
    \centering
        \includegraphics[scale=0.45]{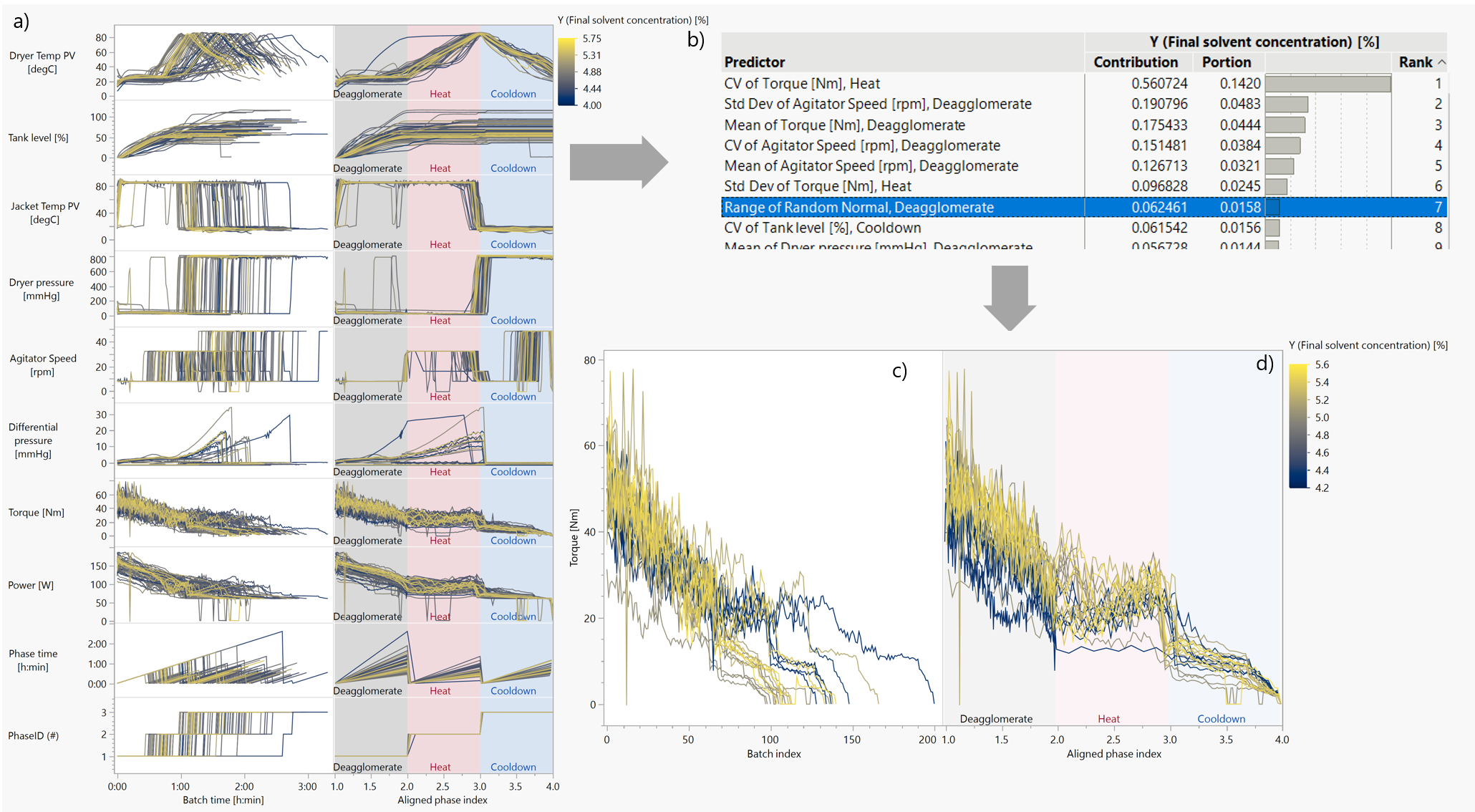}
    \caption{AutoML approach for batch process data. Statistics of all the sensors (tags) per phase and batch are calculated (a). Then a random forest model (predictor screening in b) is used to identify which of these statistic summaries (called landmarks or fingerprints) has the strongest correlation (contribution) towards the target (solvent concentration, in this case). Random noise was introduced in the analysis as a synthetic tag so it can be used as a signal-to-noise threshold (highlight contribution in b). In this example, the sensor measuring the torque (c and d) shows a pattern where high and low torque trajectories correspond to high and low solvent concentration). }
    \label{fig:Pred_Explainer}
\end{figure}

The results of this approach are shown in Fig. \ref{fig:Pred_Explainer}. First multiple statistics (e.g. mean, max., min., standard deviation, coefficient of variance) for every sensor, phase, and batch are calculated. Then a random forest model (called Predictor Screening in JMP [SAS Inst.]) is used to list the contribution of each predictor. Random noise was added as a fictitious tag, so it is used as a threshold for variable selection\cite{Wu2007Mar}.
This simple example illustrates how using only a data-driven analysis we went from multiple tags (sensor data) to only one. As shown in Fig. \ref{fig:Pred_Explainer}c ---and perhaps more clearly in Fig. \ref{fig:Pred_Explainer}d thanks to the alignment--- high and low values of solvent content correspond to high and low torque values in the deagglomerate and heat phases. A discussion on the difficulty of interpreting these results follows.

\paragraph{ExplainableAI (model interpretation)}

Industrial data is messy, there are always multiple special causes that modify the process ---sometimes intentionally as part of continuous improvement initiatives, many others unintentionally. The screening approach shown above uses summary statistics to reduce the number of information to look at. Naturaly, other methods for variable selection could be combined with this approach.\cite{LU201490_var_selection_PLS}. However, once the tags (sensor data) are reduced, a process expert will need the full trajectory (trend) to interpret the validity of the insights found\cite{Mowbray2022_ML_review} (e.g. is it a cause or a consequence that can be used to build a soft-sensor?). 

Advanced techniques for model interpretation can be applied to summary data (e.g. SHAP \cite{NIPS2017_7062,wang2021shapley,Mowbray2022_ML_review}). However, for batch process data a natural step will be to analyze the subset of tags using FPCA. 

If DTW is used to pre-align the data (see Annex \ref{app:batch data alignment}), correlation analysis can be done using the time-warped variable as a target to explain what operating conditions make a batch fast or slow compared to the selected batch reference. \cite{Spooner2018a,  Gonzalez-Martinez2011, Spooner2018b}.

\subsection{Anomaly detection for batch processes (unsupervised learning)}

\begin{figure}[!htb]%[!htb]
    \centering
        \includegraphics[scale=0.6]{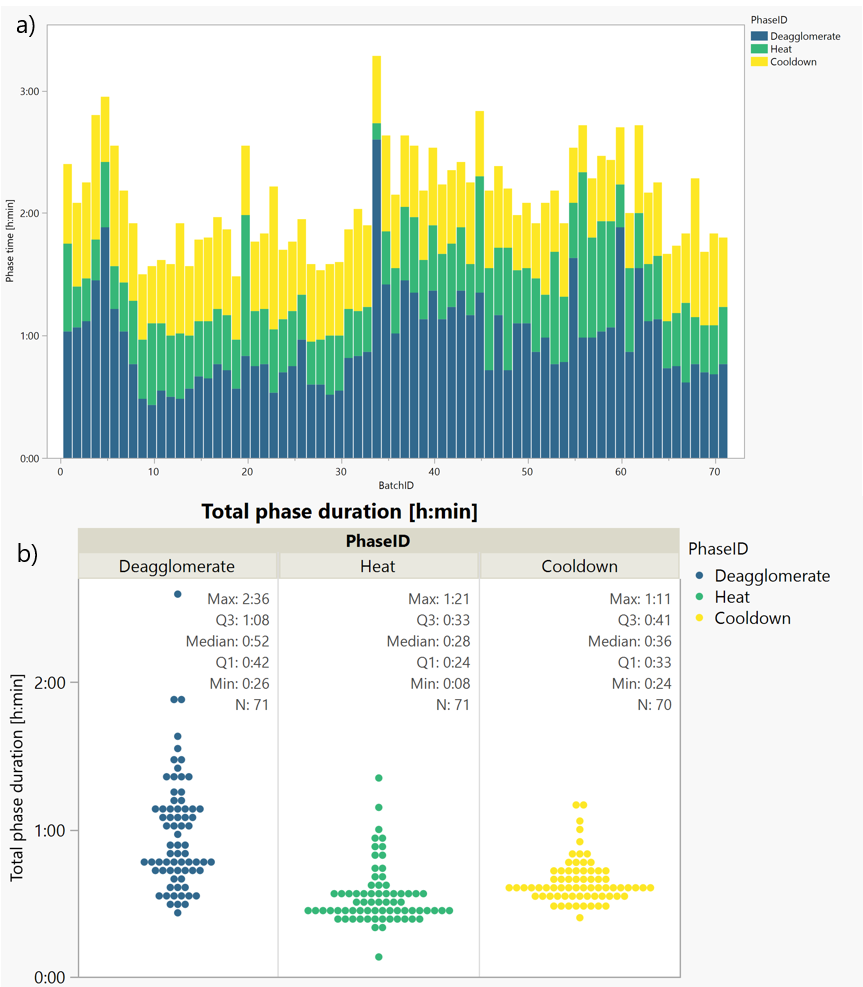}
    \caption{KPI monitoring of batch duration (a) and its accumulated variability (b)}
    \label{fig:duration_monitoring}
\end{figure}

Batch-to-batch key process indicators (KPIs) are used in the industry to monitor variability we want to control and further investigate when needed. These are, for example, quality or productivity parameters such as batch and phase duration (see Fig. \ref{fig:duration_monitoring}).

In machine learning terminology, when we want to detect anomalous patterns between data inputs (X's) such as these KPIs, we are framing the problem into unsupervised learning. This means we don't need to classify in advance if batches are normal or abnormal. We want to find these groups in our data without having to put 'labels' on them. Therefore, this classification analysis can be done automatically without expert input in the first place, hence it is 'unsupervised.'

For example, the dryer dataset has certain variability with respect to its phase and batch duration (see Fig. \ref{fig:duration_monitoring}). Looking only at past events, which of these batches can be considered normal operating conditions and which were out-of-control?

\subsubsection{Statistical process control}

Traditionally, control charts have been used to answer this question: are we observing a process change or not? As explained above, unsupervised machine learning is also used for the same purpose, as it can separate and cluster different patterns seen in past events.

\begin{figure}[!htb]%[!htb]
    \centering
        \includegraphics[scale=0.7]{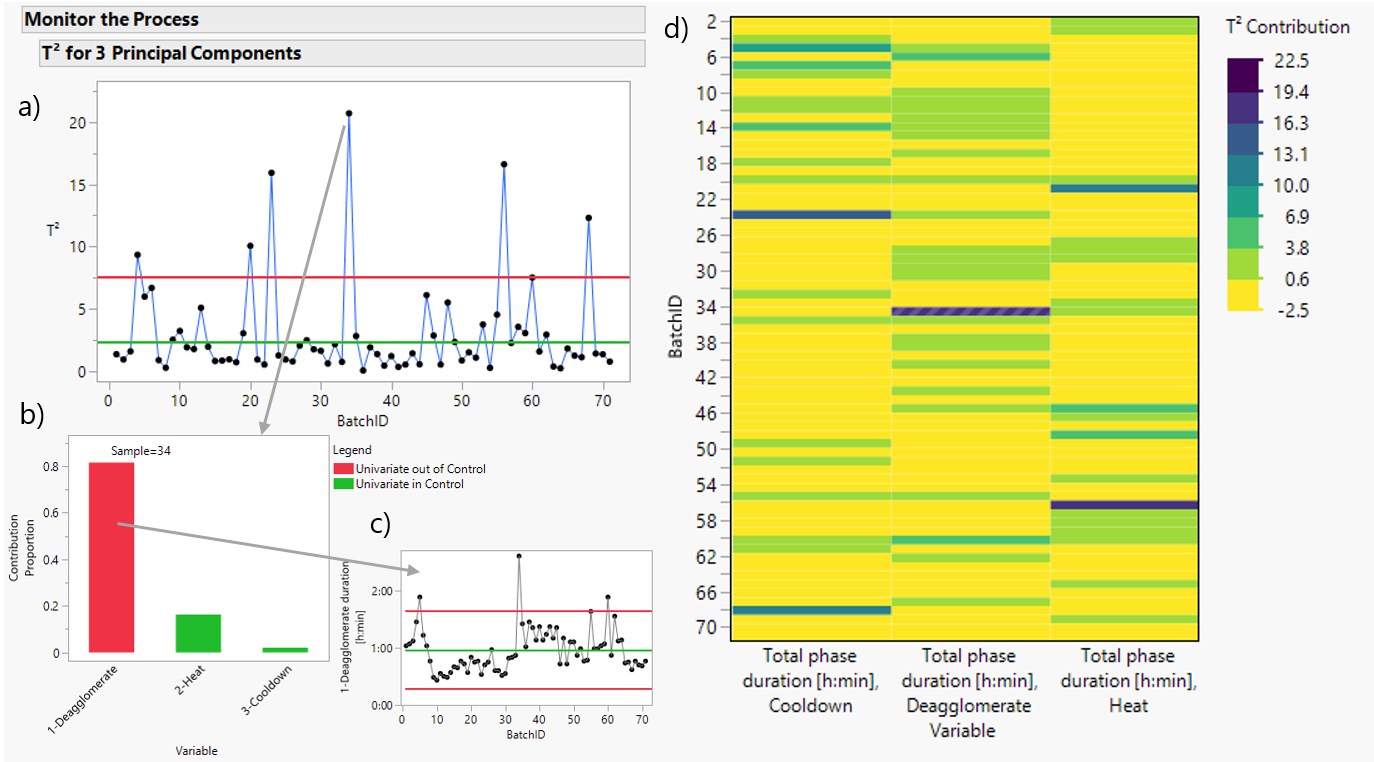}
    \caption{A multivariate control chart monitoring production KPIs (batch phase durations). The PCA-Hotelling T2 (a) calculates how different the KPIs were compared to past events. Individual contributions of deagglomeration, heat, and cooling phases can be obtained (b) and checked in univariate control charts (c) for further analysis. A heatmap with the contribution of each duration phase for all the batches can help to recognize operational changes (d).}
    \label{fig:MSPC_duration}
\end{figure}

If there are different KPIs to monitor, the correlation between them can also be monitored. This way, even when KPIs are individually under control, an anomaly can be detected when their ratio is no longer maintained.
Multivariate control charts use PCA to monitor several KPIs, even if some are redundant  (see Fig. \ref{fig:MSPC_duration}a) . For each new batch, the statistical model (called Hotelling T2) calculates how distant the new set of KPIs is different from the past. In the dryer dataset, we can see batch number 34 with a duration that is anomalous (see Fig. \ref{fig:MSPC_duration}b and c). The contribution of each KPI can be visualized in a heatmap to detect patterns during operations (Fig.  \ref{fig:MSPC_duration}d).

Multivariate Statistical Process Control (MSPC) Charts are widely used in industry to monitor industrial batch processes \cite{MacGregor1994,Wold2009,Nomikos1995,garcia_munoz_batch_2003,garcia_munoz_batch_2004, MacGregor2005May,Nomikos1994}.  Comprehensive KPIs are need to summarize important and easy to interpret information. Otherwise, batch anomalies will go unnoticed until they have visible effects on other parts of the process that are being tracked. 

\newpage

\subsubsection{Functional statistical process control}

Process engineers have limited time to create, monitor, and modify KPIs for their processes. A crucial benefit of functional statistical process control is to relieve them of this step. In fact, the intuition behind this analysis is to compare batch trajectories instead of pre-selected KPIs. This process can be automated using Functional PCA, as it can find the intrinsic trajectories for each sensor (FPCA shape functions).

\begin{figure}[!htb]
    \centering
        \includegraphics[scale=1]{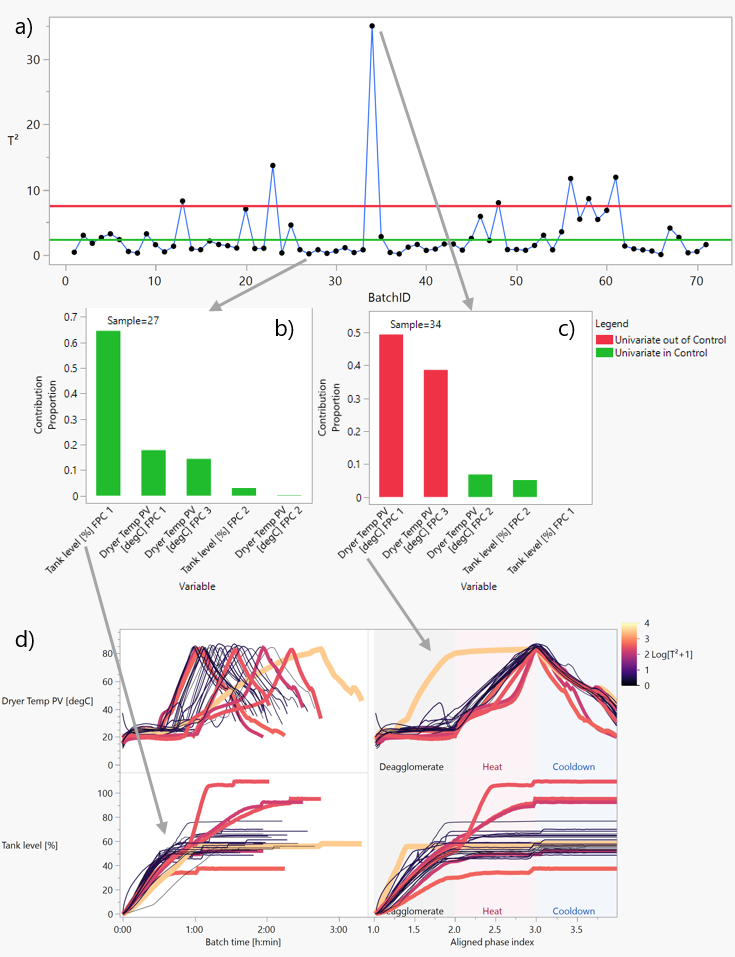}
    \caption{Multivariate control chart for batch processes using FPCA. A Hotelling-T2 control chart is built (a) monitoring scores coming from an FPCA analysis with batches that have been pre-aligned. The individual contributions for the tank level and temperature of the dryer are shown for a batch that is under control (b) and another that is out-of-control (c). Batch trajectories for these two sensors are colored (d, where a log transformation is applied to aid the visualization of normal and abnormal batches). The pre-alignment done previous FPCA helps this method to identify anomalous batches regardless their impact in duration (see aligned trends in d).}
    \label{fig:Functional_MSPC}
\end{figure}

As an example (Fig. \ref{fig:Functional_MSPC}a), a multivariate control chart can quantify how anomalous drying batches are regardless of their time duration. First, a pre-alignment of the batches is done using the phases as index. This step is important as it removes from the analysis any anomaly directly related to the batch duration ---which can be more easily tracked as shown in Fig. \ref{fig:MSPC_duration}.  Then,  functional components are used as fingerpints for both the tank level and drying temperature (Fig. \ref{fig:Functional_MSPC}b and c). The T2 score can color batch trajectories to highlight anomalous behavior (Fig. \ref{fig:Functional_MSPC}b and c). When looking at the aligned batches in Fig. \ref{fig:Functional_MSPC}d, anomalous patterns for both the temperature and tank level can be seen.

 Using FPCA, the only input from process engineers will be which tags (sensors) to monitor closely. However, if there is a specific KPI such as quality or production, an AutoML screening analysis using correlation can be performed first (see previous example looking at the solvent content). Reducing the number of tags to monitor will aid the process engineer in focusing only on abnormal behaviour that seems to affect quality and production.

\section{Challenges and Opportunities} \label{Chall_opportunities}

\subsection{OT / IT integration}
\begin{figure}[!htb]%[!htb]
    \centering
        \includegraphics[scale=0.75]{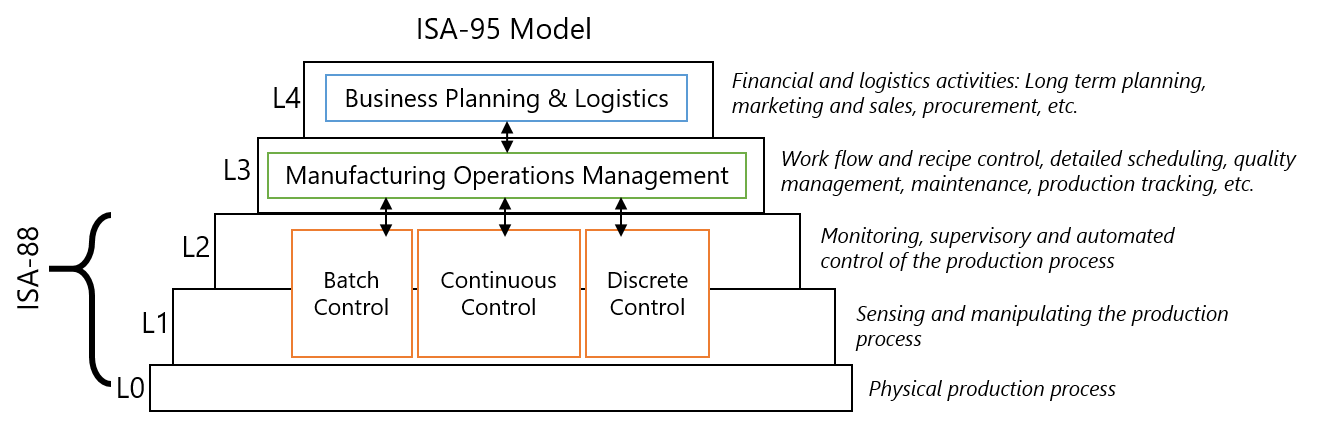}
    \caption{The standard ISA-95 interfaces between enterprise and control systems, acting as the operating framework for manufacturing. ISA-88 is focused on the process and its control. }
    \label{fig:ISA_95}
\end{figure}

In manufacturing industries, the need for handling all the information coming from sensors (called tags) and lab measurements created several standards (ISA\cite{ISA_org}, OPC\cite{OPC_foundation}, OMAC\cite{OMAC_org}, NAMUR\cite{Namur_org}). Nowadays, Operational technologies (OT) can monitor, control and optimize a plant. Although the existing standardized and industrialized solutions are robust, data is often compartmentalized in highly specialized applications (following both ISA-95\cite{Virta2010,jansen2014importance_isa95} and ISA-88\cite{hawkins2010_isa88, woodcliff_2002_isa88, OSAKA_2015_isa88}, shown in Fig.\ref{fig:ISA_95}), and only part of the information is effectively shared across. 

\begin{figure}[!htb]%[!htb]
    \centering
        \includegraphics[scale=0.65]{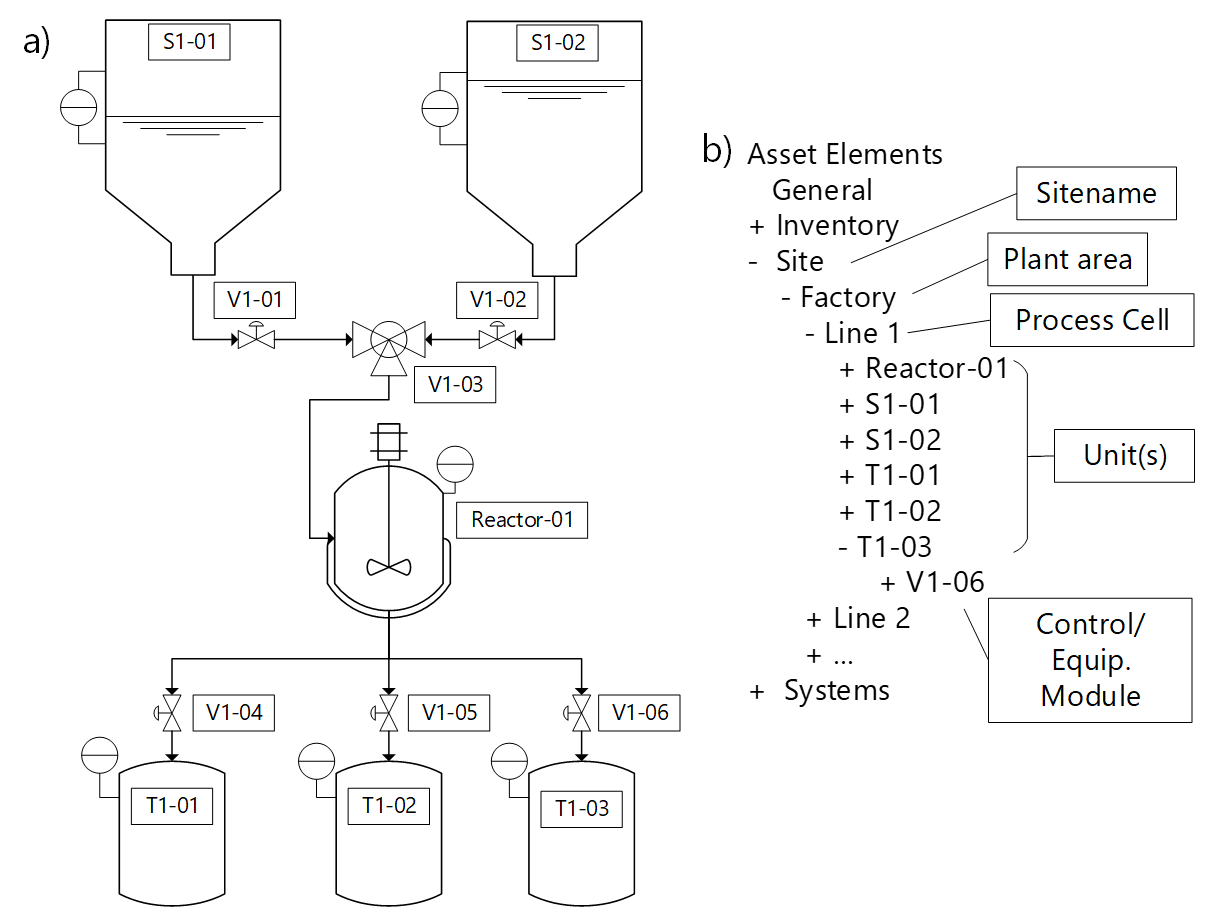}
    \caption{Simplified process (a) following ISA-88 to standardized asset hierarchy (b) in the plant. }
    \label{fig:ISA_88}
\end{figure}

\begin{figure}[!htb]%[!htb]
    \centering
        \includegraphics[scale=0.9]{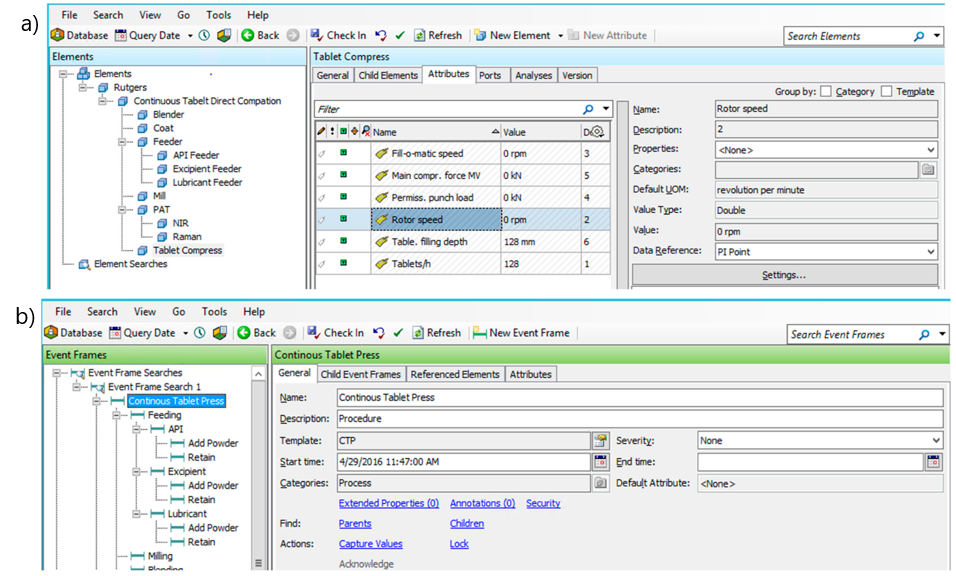}
    \caption{Osisoft PI Asset Framework (a) and Event Framework (b) from Aveva. Adapted from \cite{2018_Cao_Pharma_Manufacturing}}
    \label{fig:PI_AF_EF}
\end{figure}

ISA-88 also defines data structures (see Fig.\ref{fig:ISA_88} that can be stored and augmented in databases from Production Information Management System (PIMS)  (e.g. Fig. \ref{fig:PI_AF_EF} shows the interface for Osisoft PI Asset Framework/Event Framework from Aveva). With this contextual information one can re-apply the same analytical methods to similar units. For anomaly detection, having a map on where the anomalies occurred can also facilitate root-cause finding. 

The specialization needed to configure and maintain OT technologies (DCS, PIMS, LIMS), is also a challenge compared to the resources and facilities brought by IT technologies. Combining OT/IT technologies to avoid data duplication while supporting different business initiatives is far from trivial. This is why, the concept of scale is often referred to as handling vast amounts of data in the cloud, where more resources can be externalized. Consequently, there is a risk of creating ad-hoc solutions project by project.

The industrial applications shown in this book chapter do not require a new IT cloud architecture, but 1) specialized existing software that can natively access and facilitate the analysis of industrial batch data and 2) training of process engineers (domain experts). Improving data quality following the international standards and understanding which contextual data is stored in the different layers (ISA-95/88) will only be prioritized when a data-ownership culture is established at plant level.

\subsection{Process control vs Advanced Analytics}
As demonstrated in this book chapter, Advanced Analytics (AA) can be very useful to monitor and diagnose batch processes. It should be clear to the reader how AA uses data to facilitate these tasks to subject matter experts (SMEs). In the case of chemical engineering, the most common result obtained when applying AA techniques are very straightforward and intuitive for SMEs, e.g. a valve or a sensor was not working properly, a control scheme is not performing as it should or raw material quality has changed over time. That is why, at least in process industries, AA most of the time does not enter directly in the solution design. Very often, simpler, well-known and robust solutions are used once the root-cause is identified. The field of process control is definitely among these, starting from simple PID, DCS automation, advanced regulatory control (ARC) up to more complex solutions like Advanced Process Control (APC), Model Predictive Control (MPC)\cite{macgregor_MPC_batch,nagy2003robust,nagy2007real} and Real Time Optimization (RTO).

AA can be regularly used to monitor and further improve the already implemented process control solutions. For example, fingerprints or FPCA can be used to quickly identify where the loss in control performance is located. FPCA can also identify and classify a set of trajectories for golden batch applications (e.g., trajectory control).

Finally AA can also be used for System Identification, a topic discussed in our review\cite{Mowbray2022_ML_review}

\paragraph{Real time applications} 
Although real-time applications were not covered, a batch-to-batch analysis of past data will always be the first step. DTW can be adapted to alert of anomalies in near real-time \cite{Kassidas1998,Gonzalez-Martinez2011,Spooner2018a,Spooner2018b}, although what corrective action to apply can be unclear for an inexperienced operator. 

\newpage
\appendix

\section{Batch Data Alignment} 
\label{app:batch data alignment}

\subsection{Batch data structure}
\begin{figure}[!htb]%[h]%[!htb]
    \centering
    \includegraphics[scale=0.45]{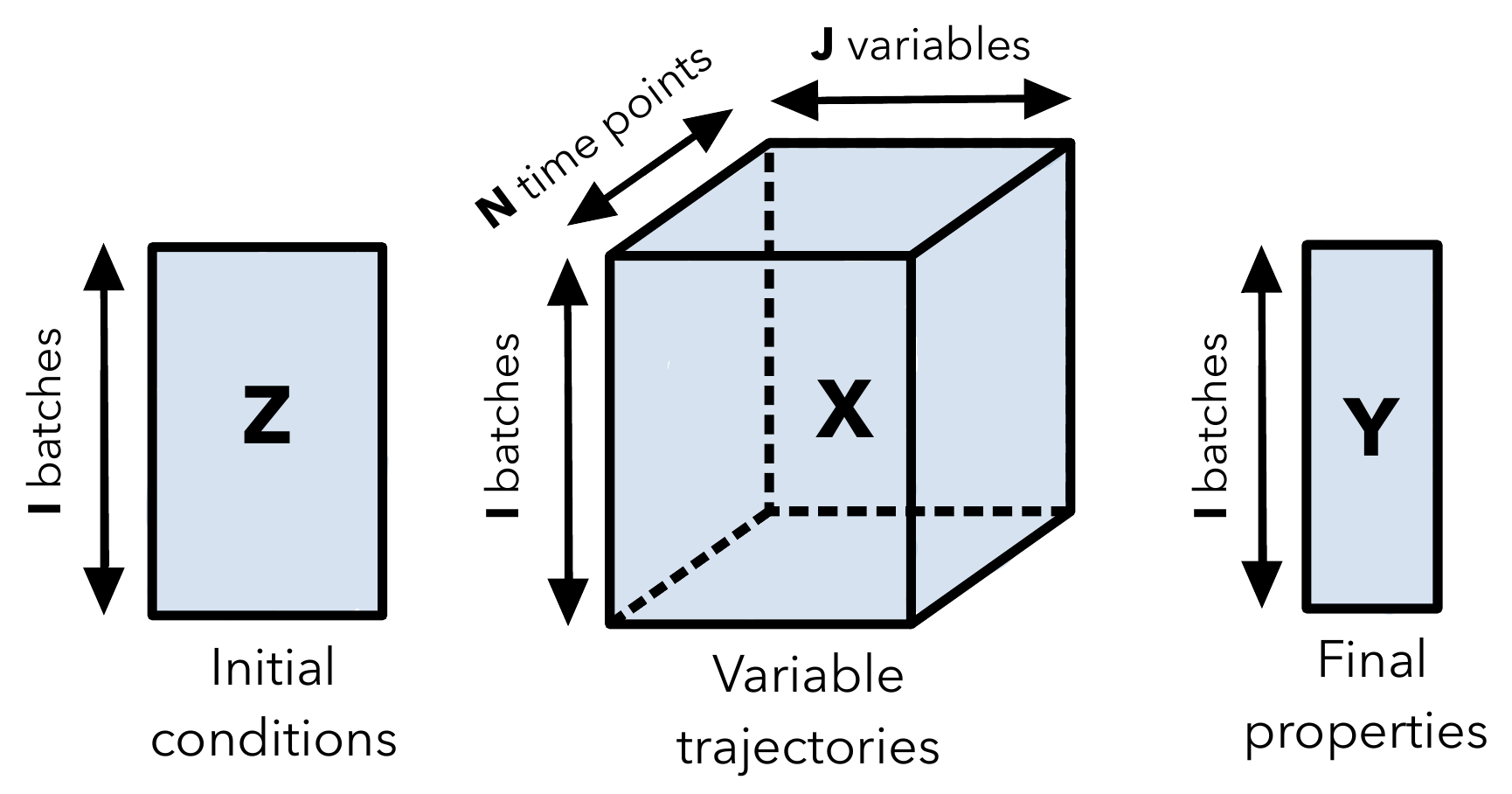}
    \caption[Batch data structure]{Structure of batch data.}
    \label{fig:batch_data_nature}
\end{figure}

Every time a batch is manufactured, three types of data are collected: (i) the initial conditions (\textbf{Z}), (ii) the variable trajectories (\textbf{X}) and (iii) the final properties (\textbf{Y}) \cite{Kourti1995,Wold2009} (Fig. \ref{fig:batch_data_nature}) . Each additional batch run constitutes an additional row in matrices \textbf{Z} and \textbf{Y}, and an additional layer in the three-dimensional tensor \textbf{X} \cite{Wold2009} with \emph{I} batches, \emph{J} sensors (also called tags) and \emph{N} time-sampled points until the end of the batch \cite{Spooner2018b}, while \textbf{Y} is typically an offline quality measure after batch completion \cite{Wold2009,garcia_munoz_batch_2003}.

\begin{figure}[!htb]%[h]%[!htb]
    \centering
    \includegraphics[scale=0.6]{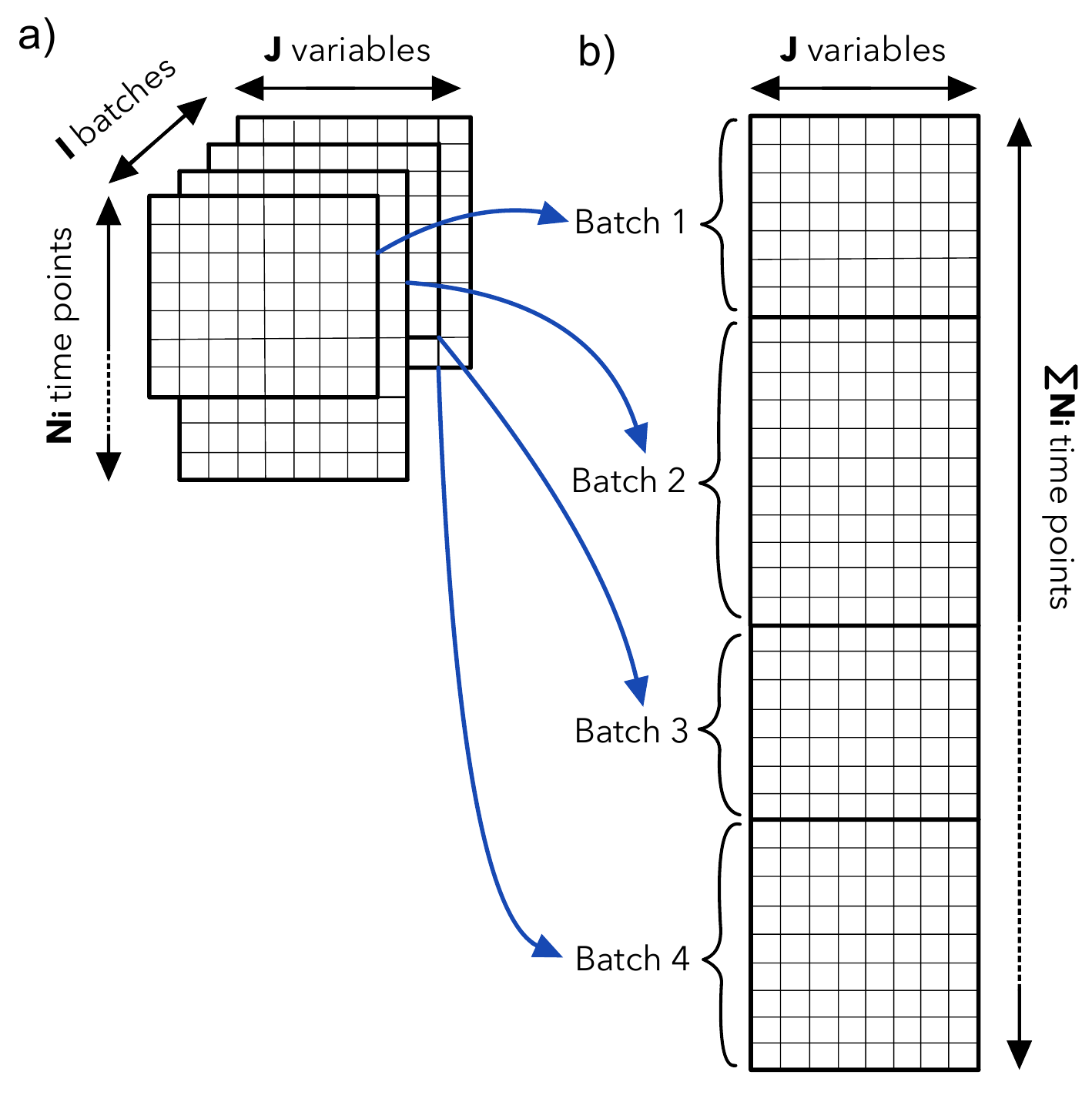}
    \caption{Batchwise unfolding of tensor \textbf{X}. a) Folded, b) unfolded. \emph{J} variables correspond to sensor readings during each batch (tags)}
    \label{fig:data_recording}
\end{figure}

When performing data analysis, it is common to work on chronologically unfolded data, also known as batchwise unfolded or BWU (Fig. \ref{fig:data_recording}), which corresponds to the structure in which historians store (batch) process data.

\subsection{Batch data alignment algorithms}
In some cases batch processes are programmed to proceed according to a defined schedule, yielding batch runs with the same duration. Nevertheless, in most cases the batch duration is a variable to be adjusted in order to achieve an end point based on some measured or inferred process variable, such as the conversion of a given reactant, resulting in batches with varying duration. Batch duration is dependent upon many parameters, such as the initial conditions and seasonal changes (e.g. in the cooling water temperature) \cite{Wold2009}.

\begin{figure}[!htb]%[h]%[!htb]
    \centering
    \includegraphics[scale=0.43]{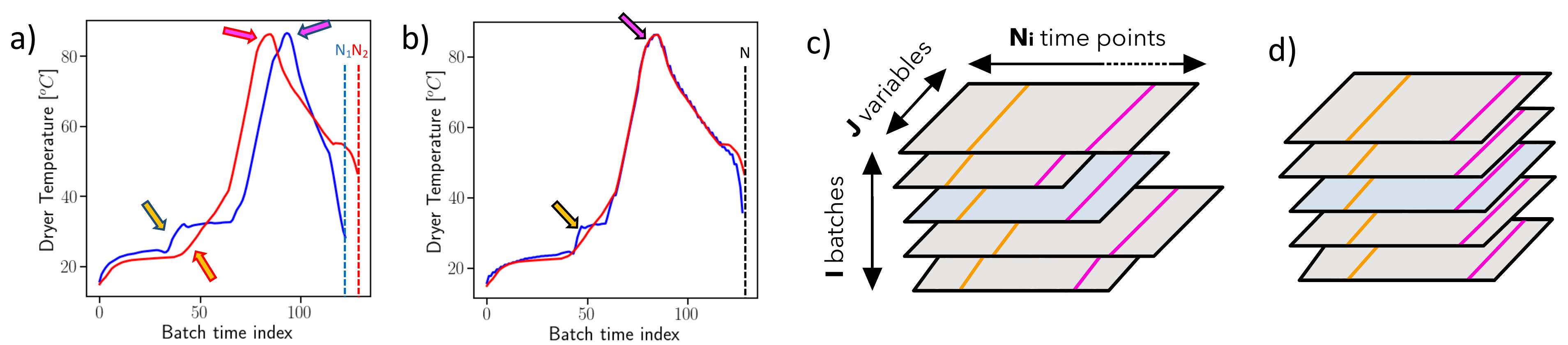}
    \caption{Two batches of different duration (a) can be aligned (b) when a reference (in red) is selected with respect the queried batch (in blue). In this case, DTW is used to assure both the same duration and alignment of features (marked with arrows). The same approach can be extended to multiple batches (c,d) such that all contain the same number of points. }
    \label{fig:X_tensor_alignment}
\end{figure}

Multivariate projection methods such as PCA and PLS require that all batches have the same number of N elements, i.e., that all batches have the same number of rows despite their duration \cite{Gonzalez-Martinez2011, Kassidas1998, MacGregor2012, Spooner2018a, Spooner2018b}. Consequently, synchronization of batch processes is a necessary step for batch process modeling. Additionally, the quality of the batch process data alignment can influence the performance of the subsequent data analysis \cite{Rendall2019}. The synchronization's goals are two-fold, namely (i) ensuring that all batches have the same number of samples and (ii) ensuring that key process events or landmarks (e.g. a step change in the temperature profile) happen at the same state of evolution \cite{Kourti2003} (Fig. \ref{fig:X_tensor_alignment}). With a set of synchronized batches, it is also easier to tell which batches deviate from the average batch trajectory, which in turn enhances the comparison between batches.

% Synchronization or alignment methods can be classified into three categories: (i) \textit{methods based on compressing/expanding the raw trajectories using linear interpolation, either in the batch time dimension or in an indicator variable dimension;} (ii) \textit{methods based on feature extraction;} (iii) \textit{methods based on stretching, compressing, and translating pieces of trajectories} \cite{Gonzalez-Martinez2011}.

Three relevant alignment algorithms are:
\begin{itemize}
\item Plotting the variable trajectories against an \textit{indicator variable} which tracks batch progress.
\item Linear alignment based on \textit{automation triggers} of the batch recipe.
\item Non linear time alignment using statistical methods that stretch, compress and translate sensor trajectories, such as \textit{Dynamic Time Warping} (DTW).
\end{itemize}

\subsubsection{Alignment based on indicator variables}
Batch process data can be aligned by plotting the variable trajectories against a measured or inferred indicator variable which correlates with batch maturity or progress (e.g. conversion). Such a variable has to be monotonically increasing or decreasing in time and show low noise levels, with the same start- and end-points across batches \cite{Wold2009, Rendall2019}. The variable trajectories have to be interpolated if they want to be plotted at equal intervals of the indicator variable, resulting in the same N time points per batch run \cite{Wold2009}. The method yields satisfactory results, being robust and, and allows online synchronization \cite{GarciaMunoz2011}. Alas, such a monotonically increasing indicator variable is not often available \cite{Kassidas1998}. For batch processes with several automation steps a different indicator variable may be available per phase, in which case trajectory alignment is performed separately \cite{GarciaMunoz2011}.

\subsubsection{Alignment based on automation triggers}
Landmarks such as peaks and valleys capture essential information in \textbf{X} \cite{Wold2009}. Such landmarks usually appear on phase changes, as an automation trigger causes a control action (e.g. opening a valve) generating major perturbations to the recorded variable values \cite{GarciaMunoz2011}. The batch trajectories can then be warped linearly phase-wise \cite{Gonzalez-Martinez2011}. Even though more sophisticated procedures exist, hyper-parameter tuning is required \cite{Brunner2021}, hence resorting to automation triggers is preferred\cite{Gonzalez-Martinez2011}. 

% This approach, which is also known as curve registration, has proven to yield smoother variable trajectories after alignment than DTW \cite{Brunner2021}.

% This approach can only be used when there are easily extractable landmark features; such a task can be challenging or even impossible for trajectories varying smoothly over time. "Another potential problem is that some landmarks may disappear in abnormal batches" \cite{Wold2009}. More sophisticated procedures are briefly discussed in Gonzalez-Martinez \emph{et al.} \cite{Gonzalez-Martinez2011}. 

%FROM BRUNNER et. al. (2021)
%In summary, curve registration techniques allow not only the alignment of variable lengths—as with an indicator variable—but also the alignment of curve features. Scenario C in Figure 3 can therefore be achieved. Since the features of many process variables occur simultaneously at phase transitions, curve registration techniques are particularly suitable for multiphase processes (Ündey and Çınar, 2002). However, applications of curve registration for bioprocesses are rare. The existence of this niche in the field of bioprocesses can at most be explained by the circumstance that the indicator variable technique is more intuitive and comparatively easy to implement and DTW can be used with less fine tuning.

\subsubsection{Dynamic Time Warping}
There are two main methodologies based on stretching, compressing and translating signals, namely (i) Correlation Optimized Warping (COW) and (ii) Dynamic Time Warping (DTW) \cite{Zhang2013}. In this section only DTW will be considered due to its superior properties, which include its online implementability as opposed to COW \cite{Zhang2013}.

%"The COW approach was at first developed to correct peak shifts in chromatographic profiles". COW stretches and expands spectral batch trajectories piecewise to synchronize the signal against a reference trajectory based on two input parameters. These are the segment length, i.e., the length of each piece, and the slack, i.e., the maximum allowed adjustment for the length of the pieces \cite{Niels1998,Qiu2021}. The optimal path is found by maximizing the correlation factors between signal pieces \cite{Gonzalez-Martinez2011}. COW shows a series of disadvantages that discourage its use for unequal length batch trajectory synchronization; namely, the algorithm shows high computational costs brought about by high complexity and high parameter sensitivity \cite{Gonzalez-Martinez2011,Qiu2021}. Furthermore, online applications are hindered by the fact that COW cannot synchronize until a block of observation is available, adding time delays to the monitoring process \cite{Zhang2013}. An application of COW to batch process monitoring can be found in Frannson \(\&\) Folestad \cite{Fransson2006}.

DTW searches for similar characteristics of two sets of variable trajectories based on a dynamic optimization scheme. The trajectories are stretched and compressed so as to make them of equal length, while preserving the original data characteristics and aligning the landmark features \cite{Qiu2021}. Additionally, DTW allows for trajectory synchronization online every time new variable measurements are available \cite{Zhang2013}.

\subsubsection{Comparison}
\begin{figure}[!htb]%[h]%[!htb]
    \centering
    \includegraphics[scale=0.42]{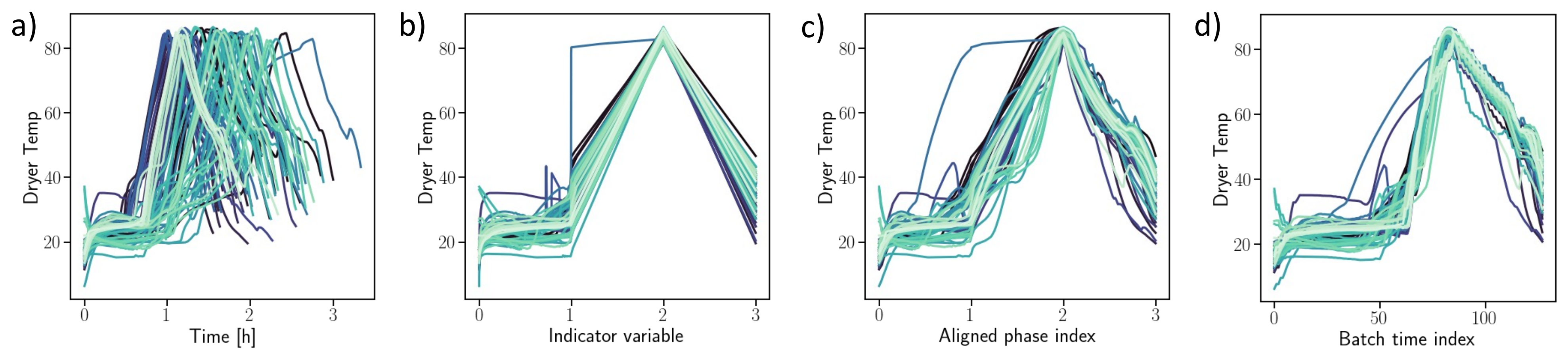}
    \caption{Visual comparison of alignment algorithms with batch trajectories colored by date, where lighter colors indicate recent time. a) Unaligned trajectories; b) Trajectories aligned by indicator variable approach (alignment variables: 1st tank level, 2nd \& 3rd dryer temperature); c) Trajectories aligned by linear interpolation between automation triggers (phases can be given equal length or the median duration of each stage); d) Trajectories aligned by multivariate DTW.}
    \label{fig:alignment_algorithm_comparison}
\end{figure}

A visual comparison of the alignment results of the batch drying case study is shown in Fig. \ref{fig:alignment_algorithm_comparison}. Additionally, Table \ref{tab:comparison_of_alignment_algorithms} shows the main advantages and disadvantages of each alignment algorithm.

In short, linearly interpolating between automation triggers is the least time consuming method, and has provided good results in the past \cite{GarciaMunoz2011}. However, it can only be applied offline, once the batch run is over (\textit{a posteriori}). For online applications such as mid-batch fault detection and soft sensor development for final batch quality prediction, the Indicator Variable approach or DTW should be employed. As mentioned earlier, the indicator variable is the preferred option due to its simplicity, robustness, and improved acceptance by the operators running the batch \cite{GarciaMunoz2011}. Unfortunately, an indicator variable with the desired characteristics is not always available, in which case DTW is an attractive alternative \cite{Rendall2019}. Additionally, different algorithms can be combined; for example, in a batch with automation triggers recorded, and with an indicator variable for the reaction (e.g. total amount of monomer fed), while the other phases can be aligned either by linear interpolation or by DTW \cite{garcia_munoz_batch_2003}.

% For the indicator variable alignment, each stage was assumed to correspond 1/3 of the batch progress, and inside of each stage an alignment with an indicator variable was performed (so both info of automation triggers and stage progress were used). The variables used were: 1st tank level, 2nd Dryer temperature, 3rd dryer temperature. Since the dryer temperature was the indicator variable, the warped trajectory is linear w.r.t. it.

\begin{table}[!htb]
	\caption{(Dis)advantages of alignment algorithms.}
    \label{tab:comparison_of_alignment_algorithms}
    \centering
    \begin{adjustbox}{width=\textwidth}
    \begin{tabular}{P{.16\textwidth}P{.28\textwidth}P{.28\textwidth}P{.28\textwidth}}
    \hline
        ~ & Indicator variable & Automation trigger & DTW \\ \hline
       Advantages & No hyper-parameter tuning, easy to apply, can be implemented online, low computational load & No hyper-parameter tuning, easy to apply, low computational load & Data-driven method, can be implemented online, alignment of minor landmarks ensured \\ 
       Disadvantages & Knowledge of process required, presence of suitable indicator variable required & Record of automation triggers required, batch-to-batch only (not online implementable) & Hyper-parameter tuning required \\ \hline
    \end{tabular}
    \end{adjustbox}
\end{table}

% DTW advantages: i) online, ii) more detail: e.g. when clustering, using automation triggers yields piecewise linear warping paths (much detail lost), while the DTW optimal warping paths hold much more info.

\subsection{Dynamic Time Warping algorithms}

\begin{figure}[h!] 
  \centering
  \includegraphics[width=5cm]{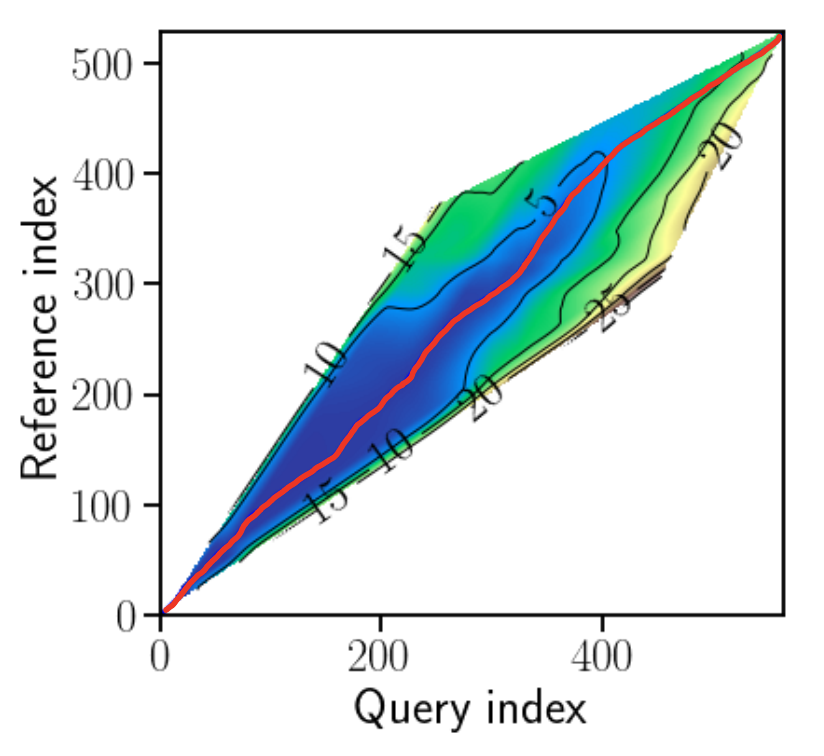}
  \caption[Optimal warping path]{Density plot of the cost matrix with optimal warping path shown in red. Notice how the optimal warping path follows the "valley" in the cost matrix, meaning that the cumulative Euclidean distance is minimized. Figure made with the \textit{dtw-python} library \cite{R_dtw_package}.}
  \label{fig:density plot with optimal warping path}
\end{figure}

DTW constructs a grid between a reference batch and a new (query) batch to be aligned. The grid, also known as cost matrix, is populated with the Euclidean distances of all combinations of the batch variable measurements. Next, a dynamic optimization scheme is followed, wherein the time index of the query batch is warped (stretched or expanded) to minimize the cumulative Euclidean distance between both batches, resulting in the optimal warping path (Fig. \ref{fig:density plot with optimal warping path}). At every iteration of the dynamic optimization, the preceding cost matrix grid cell with the lowest value is chosen, until the startpoint of both batches is reached, in a process called \textit{backtracking}. Both batches will then share the same time index, i.e. that of the reference batch. In order to guide and to ensure a realistic warping, constraints are added to the dynamic optimization problem; these are, namely, i) a global constraint, which restricts the possible warping path around the diagonal of the cost matrix and which can reduce computational load, ii) a local constraint, which restricts the possible grid cells that can be reached by backtracking at every iteration step, thus avoiding extreme time warping, and iii) boundary constraints, which ensure that the start- and end-point of the batches are aligned. The endpoint constraint can be lifted to implement DTW online \cite{Kassidas1998}.

The most common choice for the reference batch is the normal operating condition batch with median duration. The rest of the batches are warped with respect to the reference batch, obtaining a set of warped variable trajectories with the same time index as the reference batch. The reader is referred to previous literature \cite{Kassidas1998,Ramaker2003,Spooner2017,Spooner2018a,Spooner2018b,Qiu2021} for an in-depth explanations of the mathematics behind the DTW algorithm and examples of applications.

\begin{figure}[h!] 
  \centering
  \includegraphics[width=15cm]{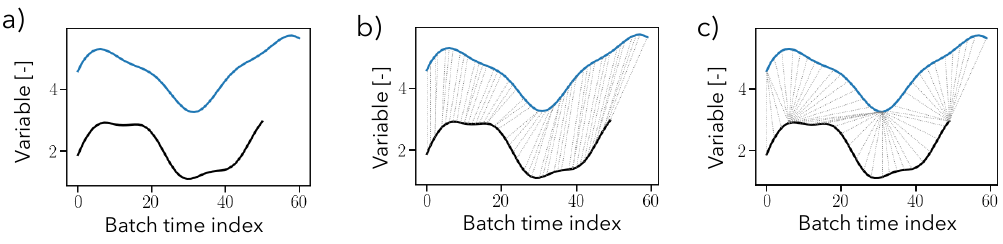}
  \caption[Presence of singularities]{a) Two synthetic curves; b) alignment without singularities; c) alignment with singularities. Singularities cause an unintuitive alignment.}
  \label{fig:singularities}
\end{figure}

The main obstacle in the application of DTW is the presence of singularities, which occur when a single point on a variable trajectory maps onto a large subsection of the other time series (Fig. \ref{fig:singularities}). Multiple variants of the original DTW algorithm proposed by Kassidas \textit{et al.} \cite{Kassidas1998} have sprung ever since, many of them addressing the formation of singularities. In the following, the different variants will be explained and classified based on the modifications made.

\subsubsection{Constraints}

\begin{figure}[!htb]%[h]%[!htb]
    \centering
    \includegraphics[scale=0.5]{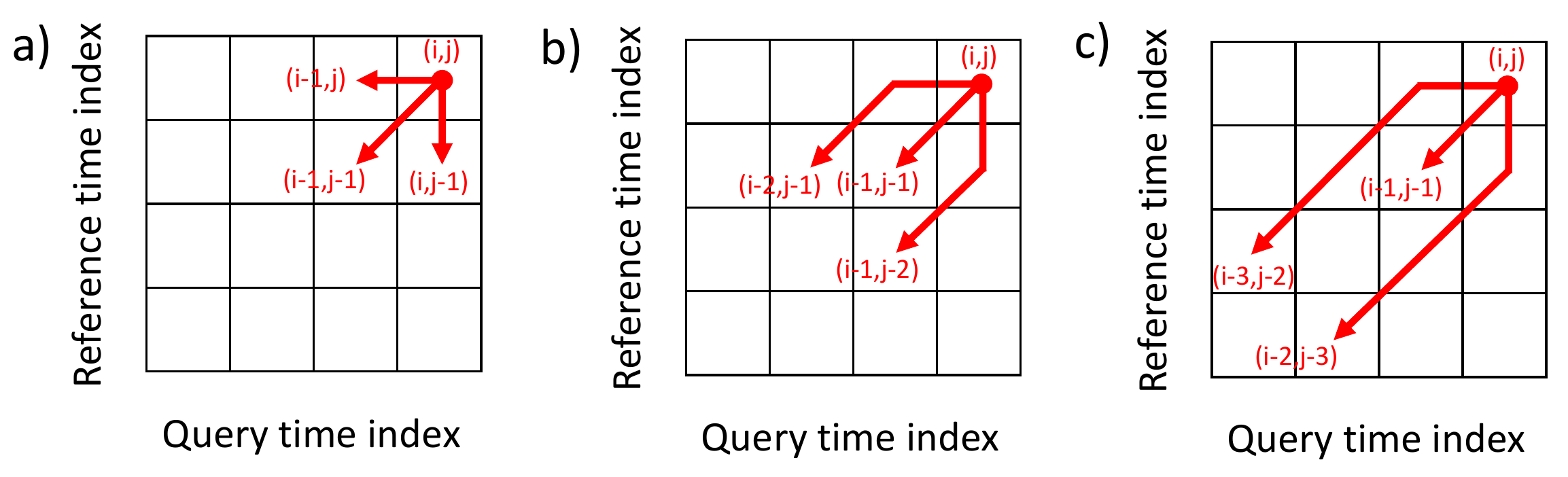}
    \caption{Examples of Sakoe-Chiba local constraints. a) P=0, b) P=1, c) P=2.}
    \label{fig:local constraint examples}
\end{figure}

\paragraph{Local constraint}
Local constraints restrict the possible steps that can be taken via backtracking in the dynamic optimization problem \cite{Kassidas1998}. The most common local constraint for batch process data alignment is the one proposed by Sakoe-Chiba \cite{SakoeChiba}, although more alternatives exist \cite{Rabiner1993,Myers1980}. Sakoe-Chiba local constraints define the number of diagonal steps that are allowed after a vertical or horizontal step in the cost matrix, defined by P (Fig. \ref{fig:local constraint examples}). 

\begin{figure}[!htb]%[h]%[!htb]
    \centering
    \includegraphics[scale=0.5]{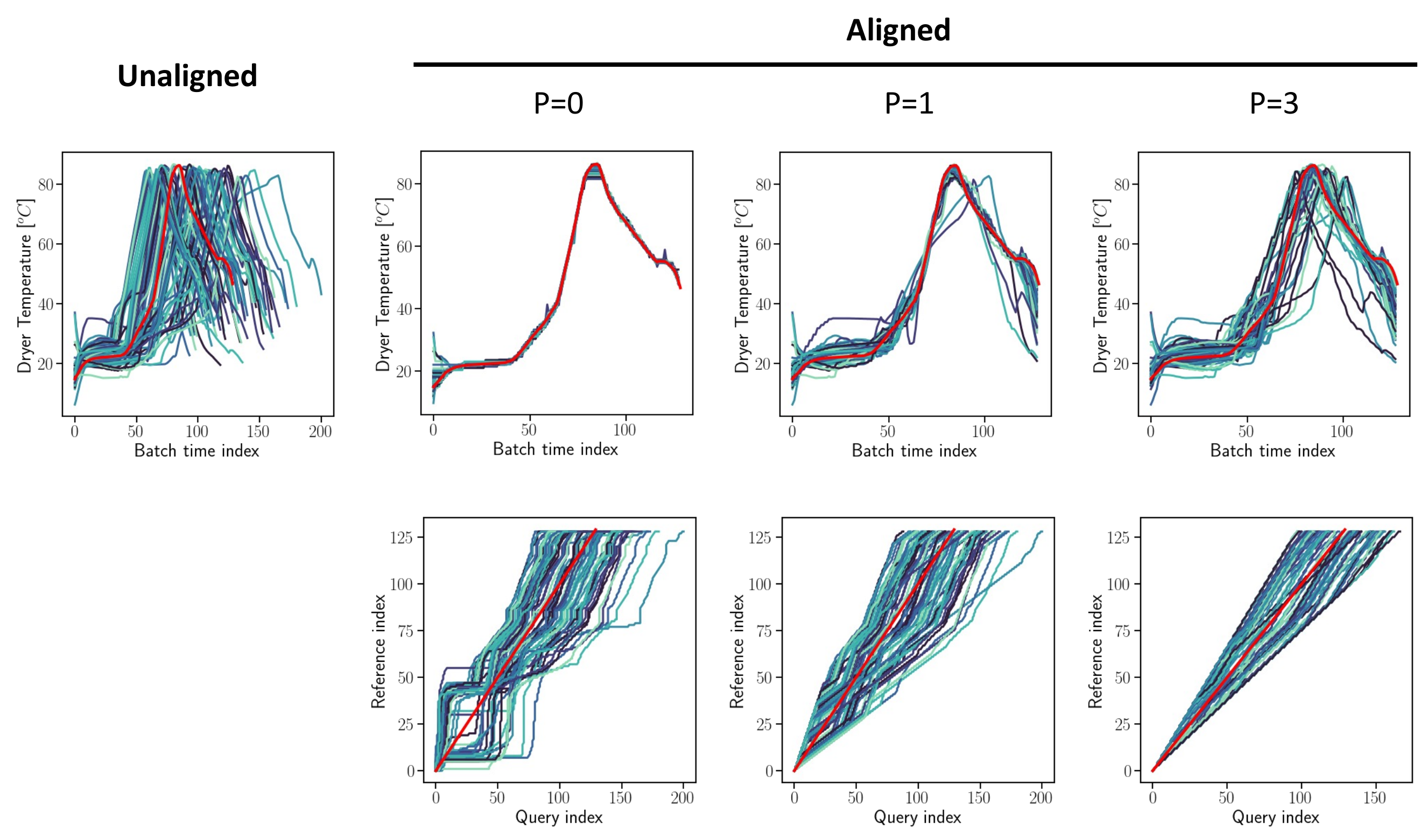}
    \caption{Sakoe-Chiba local constraint values (P) modify the results of DTW. Batch reference (red) is the one with median duration, that is also on-spec (colormap indicates batch date). The upper row shows the obtained alignment, while the lower row shows the respective optimal warping paths. When optimization is left unconstrained (P=0), singularities appear. As the constraint increases, the alignment tends to perform a linear interpolation.}
    \label{fig:local_constraint_P}
\end{figure}

\begin{figure}[!htb]%[h]%[!htb]
    \centering
    \includegraphics[scale=0.52]{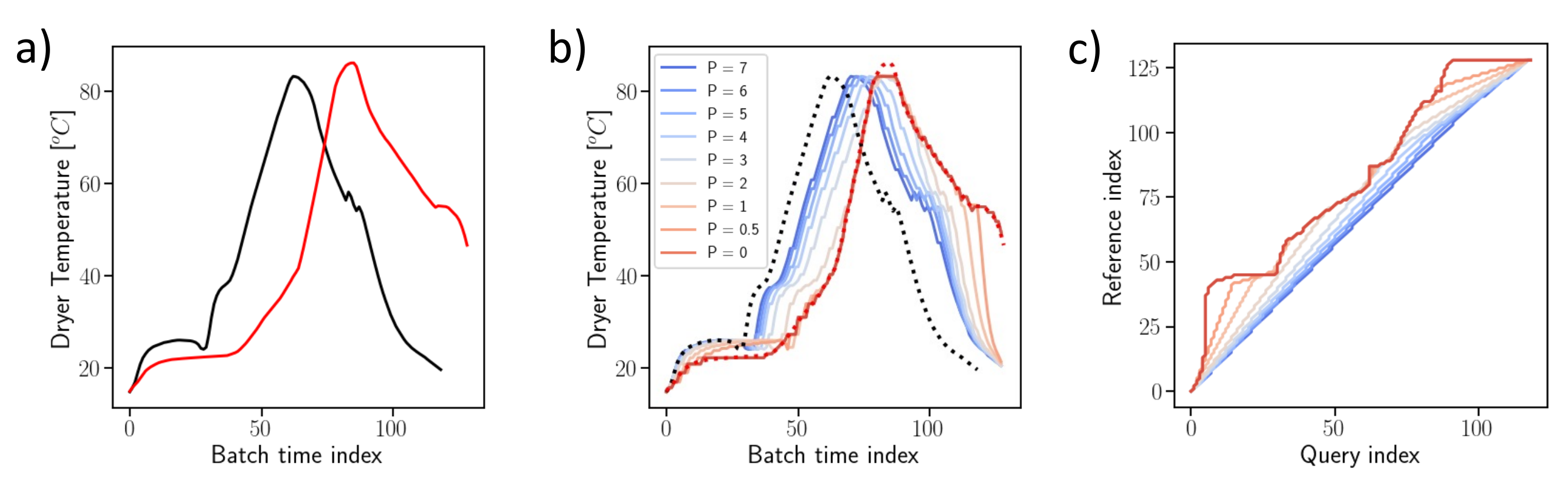}
    \caption{Effect of Sakoe-Chiba local constraint P selection. a) Unaligned trajectories (reference in red and query batch in black); b) aligned trajectories; c) optimal warping paths. As P approaches 0, more aggressive time distortion is allowed. On the other hand, as P increases, any local stretching is penalized, meaning that only a linear interpolation is performed between the start and the end of the batch with respect to the reference.}
    \label{fig:local_constraint_P_batch_13}
\end{figure}

When P approaches 0, DTW is granted full freedom to warp any time point as it deems to obtain a close match between the reference and the warped query, favoring the presence of singularities. On the other hand, as P increases, with the limit being the absolute difference between the number of time index points of the batches, the warping tends to approach a linear interpolation from beginning to end. In the middle, there is some intermediate P value with a desirable tradeoff between reduced time distortion and an acceptable alignment (Fig. \ref{fig:local_constraint_P}). The same logic applies to the alignment of a single batch (Fig. \ref{fig:local_constraint_P_batch_13}). A possible way to calculate the optimal P value is presented in Spooner \textit{et al.} \cite{Spooner2017}.

% Since there is great variation in batch completion time, relatively low P values are required. If the variation would be lower, the sweet spot of the local constraint would be at a higher P value. Reference in red; As P increases, it approaches a linear interpolation from the start to the end of the batch (linear warping paths). As P goes to zero, the warped query approaches the reference.

% As P approaches 0, more aggressive time distortion is allowed/favored and the warped trajectory approaches that of the reference. On the other hand, as P increases, the optimal warping path tends towards a straight line, meaning that a linear interpolation is performed between the start and the end of the batch.
% Batch 49 was chosen for the figure as it sometimes go faster than the reference, and sometimes slower

\begin{figure}[h!] 
  \centering
  \includegraphics[width=5.25cm]{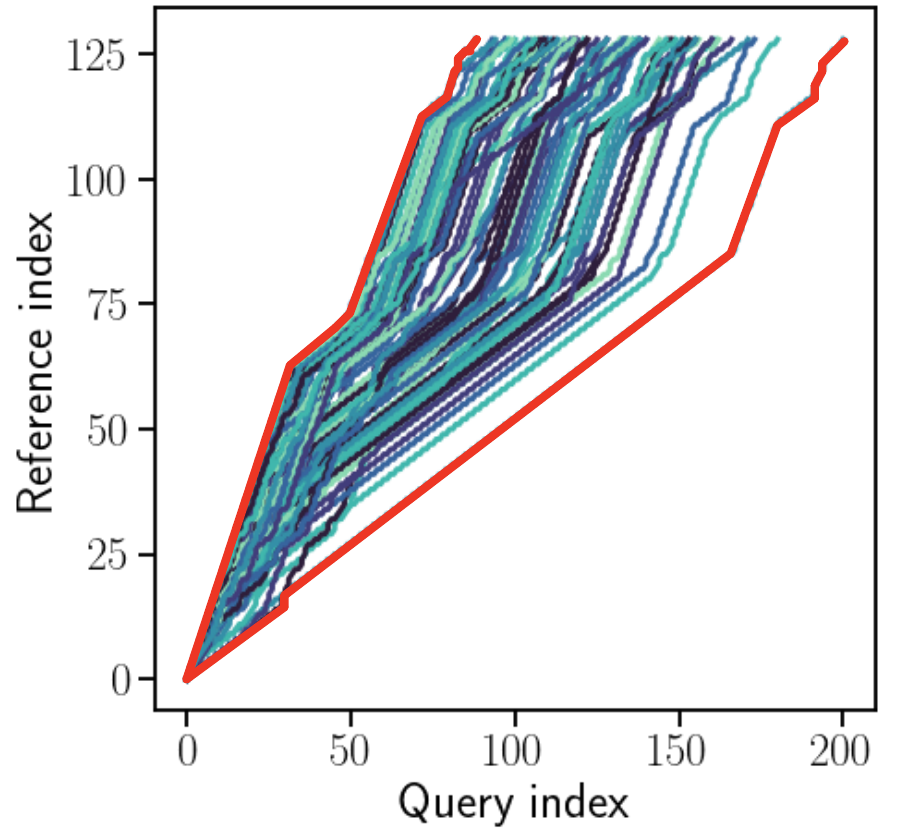}
  \caption{Upper and lower bounds of global constraint (shown in red) defined by maximum and minimum values of previous warping paths \cite{Gonzalez-Martinez2011}.}
  \label{fig:rgDTW global constraint}
\end{figure}

\begin{figure}[h!] 
  \centering
  \includegraphics[width=7.7cm]{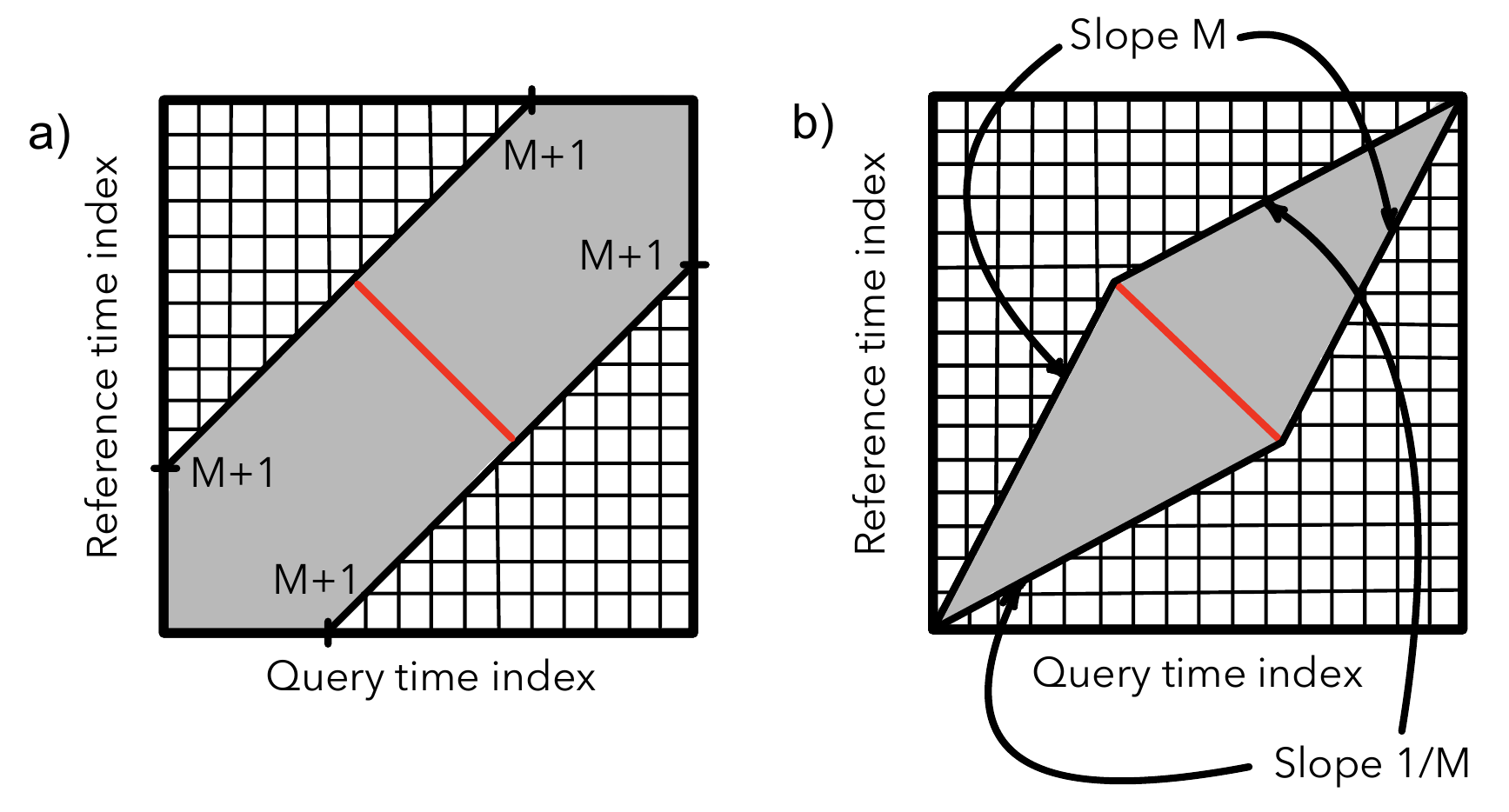}
  \caption[Examples of global constraints for DTW]{Examples of global constraints for DTW. a) Sakoe-Chiba \cite{SakoeChiba} global constraint, b) Itakura global constraint \cite{Itakura1975}.}
  \label{fig:global constraints}
\end{figure}

\paragraph{Global constraint}
The main functions of the global constraint are to avoid extreme warpings and to reduce the computational load by reducing the number Euclidean distance calculations. In the original DTW algorithm, arbitrary global constraints such as the ones shown in Fig. \ref{fig:global constraints} were proposed \cite{Kassidas1998}. Relaxed Greedy DTW (rgDTW), on the other hand, proposed setting the upper and lower bounds of the global constraint by taking the maximum and minimum grid cells reached by the optimal warping paths of a set of previous batch runs (Fig. \ref{fig:rgDTW global constraint}) \cite{Gonzalez-Martinez2011}. Its main advantages are that the computational load is reduced for online use, and an improved alarm rate when used for fault detection, as it reduces the variability in the online warping function every time a new measurement is taken by performing DTW on a moving window. Yet, it does not allow a bad alignment to be revised \cite{Gonzalez-Martinez2011}.

Based on the rgDTW, the multisynchro DTW algorithm was developed to tweak the online rgDTW algorithm based on the asynchronism type involved, which is defined by the evolution of the process pace and the batch duration, e.g. by lifting the start or endpoint constraint as needed \cite{GonzalezMartinez2014}. The constrained selective DTW (csDTW) finds regions with characteristic features and selectively synchronizes them. The main advantage of this approach is that it allows for the preservation of key features that characterize the batch, while the rest of the regions are warped without much information loss \cite{Lu2016,Spooner2018a}.

\paragraph{Boundary constraints}
As mentioned earlier, online DTW is able to warp batch trajectories which end at different states of evolution of the same batch recipe \cite{Kassidas1998}. In such a way, every time a new measurement set is taken the optimal warping path is recalculated (no end-point or boundary constraint is applied).

\subsubsection{Batch trajectory selection and pre-treatment}

\paragraph{Derivative DTW variants}
In the original DTW algorithm, which will be referred to as classical DTW (cDTW) hereinafter, the original batch trajectories are used for alignment. When the reference and the query trajectories have similar shapes but sharply different absolute values, the presence of singularities is favored. cDTW tries to correct the variation of the raw values by warping the time index, thus aligning points at different batch progress states, also possibly causing singularities in the alignment \cite{Zhang2013}. A possible solution is to perform DTW based on the derivative of the batch trajectories, thus aligning based on the shape of the trajectories rather than on their absolute values \cite{Zhang2013}. Several derivative DTW variants exist, which differ in the noise-filtering algorithm employed before the derivatives are taken. Some examples are the Derivative DTW (dDTW) \cite{Keogh2001}, which filters the noise by the Mill's exponential smoothing, the Robust Derivative DTW (rdDTW) \cite{Zhang2013}, which filters the noise with a Savitzky-Golay filter, and the Hybrid Derivative DTW (hdDTW) \cite{Gins2012}, which makes a piecewise linear approximation of the original trajectories.

\begin{figure}[!htb]%[h]%[!htb]
    \centering
    \includegraphics[scale=0.45]{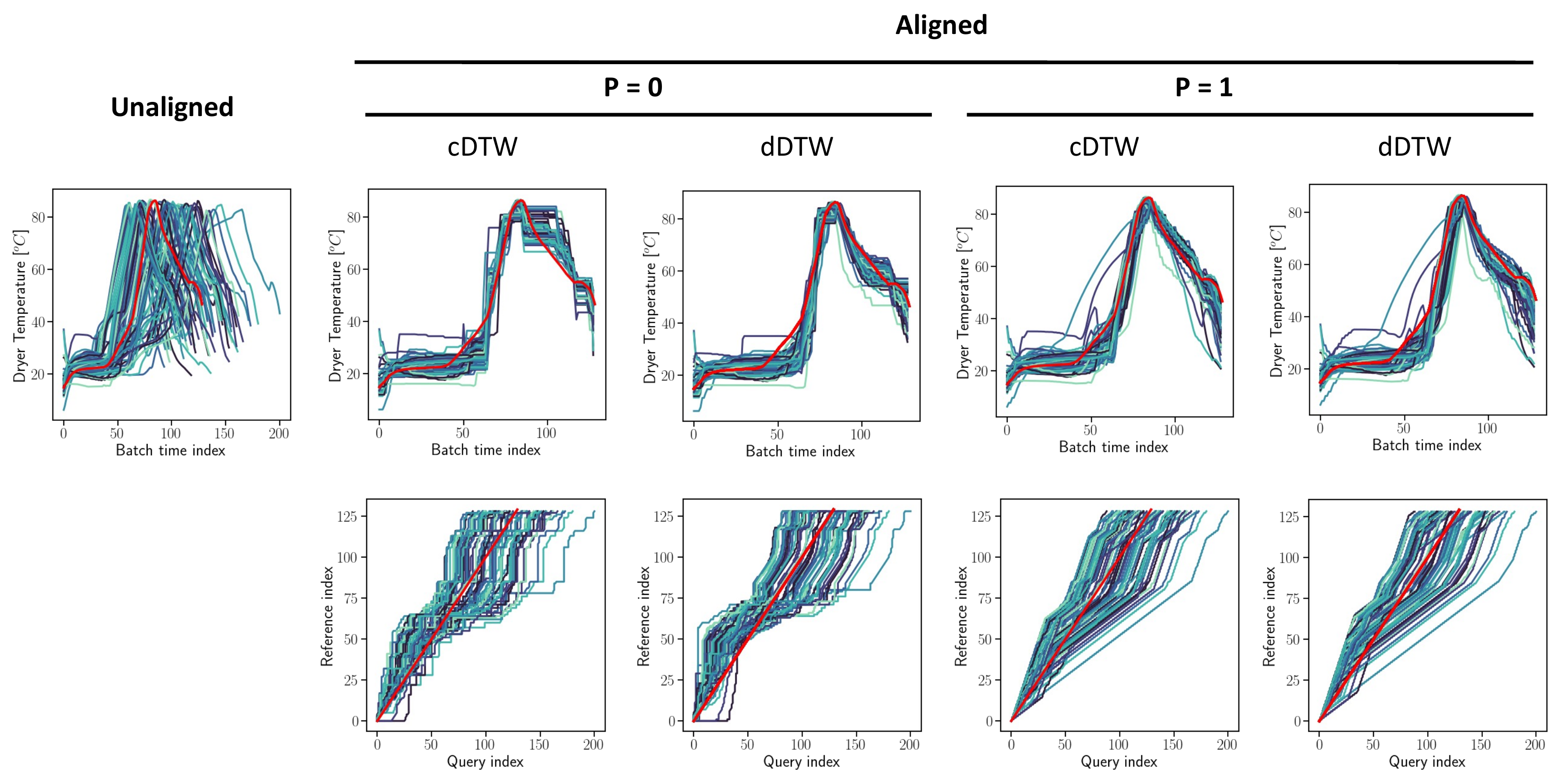}
    \caption{Combination of local constraints and variable pretreatments. If P=0, dDTW reduces the amount of singularities compared to cDTW. Similar results can be obtained when cDTW has P$>$0.}
    \label{fig:DTW_variant_and_local_constraint}
\end{figure}

If a local constraint (P$>$0) is applied, the alignment results can become largely insensitive to the variable trajectory pretreatment \cite{Arzac2022} (Fig. \ref{fig:DTW_variant_and_local_constraint}).

\begin{figure}[!htb]%[h]%[!htb]
    \centering
    \includegraphics[scale=0.76]{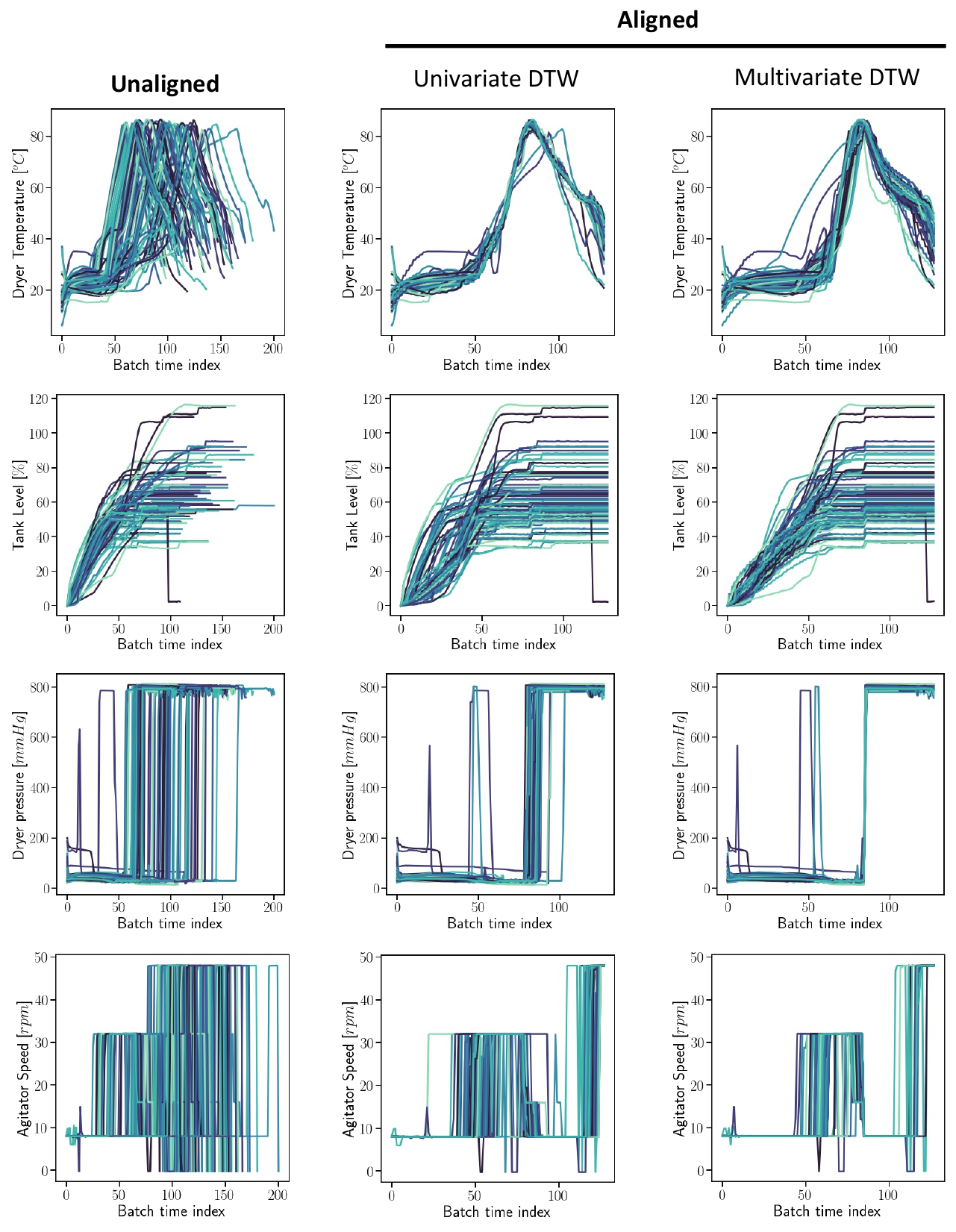}
    \caption{Comparison of univariate DTW performed using the dryer temperature only and a multivariate DTW, including the tank level, dryer pressure and agitator speed. Colormap indicates batch date.}
    \label{fig:univariate_vs_multivariate_DTW}
\end{figure}

\paragraph{Univariate and multivariate DTW}
Regarding the number of variables or tags, a single, multiple or all variables can be used for warping. Univariate DTW overfits the variable used for alignment, at the expense of the other variables' alignment, as shown in Fig. \ref{fig:univariate_vs_multivariate_DTW}. On the other hand, multivariate DTW tries to obtain the best alignment possible for all variables, making it more robust and less prone to singularities, as in the univariate case some automation triggers that hold valuable information for warping may not appear on the variable used for alignment. For the multivariate DTW, all variables can be given the same weight for warping, or preferably, the variables with the most information for warping can be prioritized \cite{Ramaker2003}. 

\subsubsection{DTW discussion}
A multivariate cDTW with variable weighting \cite{Ramaker2003} and optimal local constraint P \cite{Spooner2017} can yield the best results in terms of robustness. This involves a certain amount of hyper-parameter tuning in the first place. In case the obtained alignment is not satisfactory, then a multivariate dDTW variant may be combined with a local constraint \cite{Spooner2018a}.

\begin{figure}[!htb]%[h]%[!htb]
    \centering
    \includegraphics[scale=0.3]{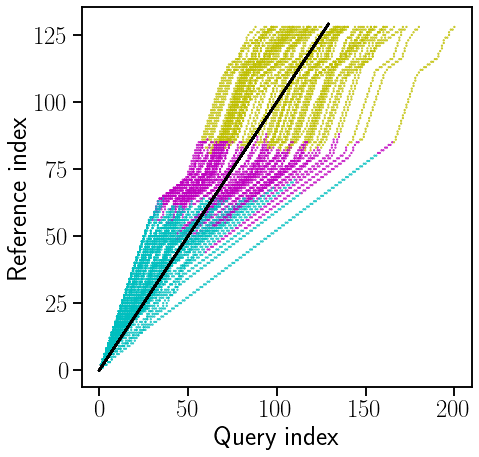}
    \caption{DTW does not necessarily align the phases (colored) since it is using a batch as reference.}
    \label{fig:DTW_stage_alignment}
\end{figure}

DTW does not necessarily align by phase (Fig. \ref{fig:DTW_stage_alignment}); rather it warps to obtain a better alignment of the variable trajectories with respect to the reference batch. If information on automation triggers is available, DTW may be performed stage-wise to align both between and within stages.

\newpage

\section*{Copyright}
This work is licensed under a Creative Commons Attribution-NonCommercial-ShareAlike 4.0 International License.

\section*{Conflict of interests}
There are no conflicts to declare. 

\section*{Disclaimer of liability}
Authors and their institutions shall not assume any liability, for any legal reason whatsoever, including, without limitation, liability for the usability, availability, completeness, and freedom from defects of the examples provided as well as for related information, configuration, and performance data and any damage caused thereby.

\bibliographystyle{unsrt}
\bibliography{bibleography}
\end{document}